\documentclass[twoside,11pt]{article}

\usepackage{subfig}
\usepackage{jmlr2e} 
\usepackage{subfig}
\usepackage{tipa}
\usepackage{mathtools}
\usepackage{fancyref}
\usepackage{float}
\floatstyle{boxed}
\restylefloat{figure}
\usepackage {amsmath}
\usepackage{amssymb}  
\usepackage{mathtools}
\usepackage{fancyhdr}
\usepackage{kantlipsum}
\usepackage{eso-pic}
\usepackage{nomencl}
\usepackage{etoolbox}
\usepackage{balance}
\usepackage{dsfont}
\usepackage[toc,page]{appendix}
\usepackage{fixmath}

\newtheorem{assumption}{Assumption}

\begin{document}

	\title{From Game-theoretic Multi-agent Log-Linear Learning to Reinforcement Learning}

\author{\name Mohammadhosein Hasanbeig \email hosein.hasanbeig@cs.ox.ac.uk \\
	\addr Computer Science Department\\
	University of Oxford\\
	Wolfson Building OX1 3QD, Oxford, UK
	\AND
	\name Lacra Pavel \email pavel@ece.utoronto.ca \\
	\addr Electrical and Computer Engineering Department\\
	University of Toronto\\
	10 King’s College Road M5S 3G4, Toronto, Canada}

\editor{}

\maketitle

\begin{abstract}%   <- trailing '%' for backward compatibility of .sty file
	%Multi-agent Systems (MASs) have found a variety of industrial applications from economics to robotics, owing to their high adaptability, scalability and applicability. However, with the increasing complexity of MASs, multi-agent control has become a challenging problem to solve. Among different approaches to deal with this complex problem, game theoretic learning recently has received researchers' attention as a possible solution. In such learning scheme, by playing a game, each agent eventually discovers a solution on its own. 
	The main focus of this paper is on enhancement of two types of game-theoretic learning algorithms: log-linear learning and reinforcement learning. The standard analysis of log-linear learning needs a highly structured environment, i.e. strong assumptions about the game from an implementation perspective. In this paper, we introduce a variant of log-linear learning that provides asymptotic guarantees while relaxing the structural assumptions to include synchronous updates and limitations in information available to the players. On the other hand, model-free reinforcement learning is able to perform even under weaker assumptions on players' knowledge about the environment and other players' strategies. We propose a reinforcement algorithm that uses a double-aggregation scheme in order to deepen players' insight about the environment and constant learning step-size which achieves a higher convergence rate. Numerical experiments are conducted to verify each algorithm's robustness and performance. 
\end{abstract}

\section{Introduction}

Most of the studies done on Multi-agent Systems (MASs) within the game theory framework are focused on a class of games called potential games (e.g. (\cite{rahili}), (\cite{yatao}), (\cite{wierman})). Potential games are an important class of games that are most suitable for optimizing and modeling large-scale decentralized systems. Both cooperative (e.g. (\cite{cp1.5})) and non-cooperative (e.g. (\cite{rahili})) games have been studied in MAS. A non-cooperative potential game is a game in which there exists competition between players, while in a cooperative potential game players collaborate. In potential games, a relevant equilibrium solution is a Nash equilibrium from which no player has any incentive to deviate unilaterally.

The concept of ``learning" in potential games is an interesting notion by which a Nash equilibrium can be reached. Learning schemes assume that players eventually learn about the environment (the space in which agents operate) and also about the behavior of other players (\cite{jair}). A well-known game theoretic learning is Log-Linear Learning (LLL), originally introduced in (\cite{blume}). LLL has received significant attention on issues ranging from analyzing convergence rates (e.g. (\cite{shah})) to the necessity of the structural requirements (e.g. (\cite{netzer})). The standard analysis of LLL relies on a number of explicit assumptions (\cite{marden}): (1) Players' utility functions establish a potential game. (2) Players update their strategies one at a time, which is referred to as \emph{asynchrony}. (3) A player is able to select any action in the action set, which is referred to as \emph{completeness} assumption. LLL guarantees that only the joint action profiles that maximize the potential function are stochastically stable. As we see in Section \ref{LiG}, the asynchrony and completeness assumptions can be relaxed separately which, results in two different algorithms called Synchronous LLL (SLLL) and Binary LLL (BLLL) (\cite{marden}) respectively. 

In this work we combine the advantages of both algorithms and relax the asynchrony and completeness assumptions at the same time. We would like to emphasize that ``synchronous learning'' used in (\cite{marden}) does not necessary mean that the whole set of agents are allowed to learn, but rather means that the learning process is carried out by a group of agents. Although, this group can be the entire set of agents, we believe that the phrase ``partial-synchronous learning'' is more accurate in reflecting what we mean by this multi-agent learning and we use this expression in the rest of this paper.

A different type of learning, originally derived from behaviorist psychology and the notion of stimulus-response, is Reinforcement Learning (RL). The main idea in RL is that players tend to use strategies that worked well in the past. In RL, players keep an ``aggregate" of their past interactions with their environment to respond in future situations and produce the most favorable outcome. Towards this direction we propose a new RL-based algorithm whose performance is much better than conventional RL methods in potential games. The following is an overview of our contributions:

\begin{itemize}
	\item The work in (\cite{marden}) studies LLL from the perspective of distributed control
	theory. Classical LLL has useful convergence guarantees, but makes several assumptions that are unrealistic from the perspective of distributed control. Namely, it is assumed that agents always act asynchronously, i.e. one at a time, and that they always have complete access to use any of their actions. Although, (\cite{marden}) demonstrates that these assumptions can be relaxed separately, we show that these relaxations can be combined, i.e. LLL can be employed without the asynchrony assumption and
	without the completeness assumption, at the same time. This, as we can see later, increases the convergence rate of the algorithm and optimizes the exploration process. Formal convergence analysis of the proposed algorithm is also presented. Note that, while in an asynchronous learning process only one player is allowed to learn at each iteration, in a partial-synchronous process a group of agents (including the whole set of agents) is able to take actions. However, in both cases, all players are aware of the process common clock.      
	\item We propose a modified Expectation Maximization (EM) algorithm that can be combined with LLL to build up a model-based LLL algorithm which further relaxes LLL's assumptions on initial knowledge of utility function. In addition to this, our modified algorithm relaxes the basic assumption of known component number of the classical EM, in order to make it more applicable to empirical examples. Through a numerical experiment, we show that by using this algorithm, both the convergence rate and the equilibrium are improved.
	\item We finally propose a model-free RL algorithm which completely drops LLL's assumptions on players' knowledge about their utility function and other players' strategies. However, as we will see later this comes with the cost of slower convergence rate. The proposed RL employs a double-aggregation scheme in order to deepen players' insight about the environment and uses constant learning step-size in order to achieve a higher convergence rate. Convergence analysis of this algorithm is presented in detail. Numerical experiments are also provided to demonstrate the proposed algorithm's improvements.
\end{itemize}
A short version of this work without proofs and generalization appears in (\cite{ifac}) and (\cite{icmlpr}). This paper discusses a number of generalizations and also proofs with necessary details. 

\section{Background}
Let $ \mathcal{G}(\mathcal{I},\mathcal{A},u) $ be a game where $ \mathcal{I} $ denotes the set of players and $ |\mathcal{I}|=N $, $ \mathcal{A}=\bigtimes_i \mathcal{A}^i$ denotes the action space, where $ \mathcal{A}^i $ is the finite set of actions of player $ i $, and $ u^i: \mathcal{A} \rightarrow \mathds{R} $ is player $ i $'s utility function. Player $ i $'s (pure) action is denoted by $ \alpha^i \in \mathcal{A}^i$, with $ \alpha^{-i}=(\alpha^1,...,\alpha^{i-1}, \alpha^{i+1},...,\alpha^N)$ denoting action profile for players other than $i$. With this notation, we may write a joint action profile $ \alpha=(\alpha^1,...,\alpha^N) \in \mathcal{A}$ as $ \alpha=(\alpha^i, \alpha^{-i}) \in \mathcal{A}$. 

In this paper, $t$ is the continuous time and $n$ is the discrete time. In a repeated version of the game $\mathcal{G}$, at each time $t$ (or at every iteration
$n$), each player $i \in \mathcal{I}$ selects an action
$\alpha^i(t)\in \mathcal{A}^i$ (or $\alpha^i(n)\in \mathcal{A}^i$) and receives a utility $u^i(\alpha)$ which, in general, is a function of
the joint action $\alpha$.
Each player $i$ chooses action $\alpha^i(t)$ (or $\alpha^i(n)$) according to the information and observations available to player $i$ up to $t$ (or iteration $n$) with the goal of maximizing its utility. Both the action selection process and the available information depend on the learning process. 

In a repeated game, a Best Response (BR) correspondence $BR(\alpha^{-i})$ is defined as the set of optimal strategies for player $i$ against the strategy profile of its opponents, i.e.
$
BR(\alpha^{-i})=\arg\!\max _{\alpha^i \in \mathcal{A}^i} u^{i}(\alpha^{i},\alpha^{-i})
$
This notion is going to be used quite often in the rest of this paper. 
\subsection{Potential Games}
The concept of a potential game, first introduced in (\cite{monderer}), is a useful tool to analyze equilibrium properties in games. In a potential game a change in each player's strategy is expressed via a player-independent function, i.e. potential function. In other words, the potential function, specifies players' global preference over the outcome of their actions.
\noindent
\begin{definition}\textbf{Potential Game}:\label{PG} A game $ \mathcal{G} $ is a potential game if there exists a potential function $ \Phi:\mathcal{A} \mapsto \mathds{R} $ such that for any agent $ i \in \mathcal{I}$, for every $\alpha^{-i} \in \mathcal{A}^{-i}$ and any $ \alpha^{i}_1,\alpha^{i}_2 \in \mathcal{A}^{i} $ we have $
	\Phi (\alpha^{i}_2,\alpha^{-i})-\Phi (\alpha^{i}_1,\alpha^{-i})=
	u^{i}(\alpha^{i}_2,\alpha^{-i})-u^{i}(\alpha^{i}_1,\alpha^{-i}),$
	where $ u^i:\mathcal{A} \mapsto \mathds{R} $ is the player $ i $'s utility function\nomenclature{$u^i$}{player $ i $'s utility function} (\cite{marden}). 
\end{definition}

From Definition \ref{PG}, when player $ i $ switches its action, the change in its utility equals the change in the potential function. This means that for all possible deviations from all action pairs, the utility function of each agent $ i $ is aligned with the potential function. Thus, in potential games, each player's utility improvement is equal to the same improvement in the potential function. 

An improvement path $\Omega$ in a potential game is defined as a sequence of action profiles $\Omega=\{\alpha_1\rightarrow \alpha_2 \rightarrow...\rightarrow \alpha_m\}$ such that in each sequence $\alpha_k \rightarrow \alpha_{k+1}$ a player $i$ makes a change in its action and receives a strictly higher utility, i.e. $u^i(\alpha_{k+1})>u^i(\alpha_k)$. An improvement path terminates at action profile $\alpha_*$ if no further improvement can be obtained. A game $\mathcal{G}$ is said to have the finite improvement property if every improvement path in $\mathcal{G}$ is finite. 

\begin{theorem}
	Every improvement path in a finite potential game is finite (\cite{monderer}).
\end{theorem}
%\textit{Proof:} Since $\mathcal{A}$ is a finite set, every improvement path $\Omega=\{\alpha_1\rightarrow \alpha_2 \rightarrow...\}$ must be finite.\\$ ~\hfill\square $\\*
This means that there exist a point $\alpha_*$ such that no player in $\mathcal{I}$ can improve its utility and the global potential function by deviating from this point. In other words
\begin{equation}
\label{nash}
u^i (\alpha_*^i,\alpha_*^{-i}) \geq u^i (\alpha^i,\alpha_*^{-i}),~~\forall \alpha^i \in \mathcal{A}^i,~~\forall i \in \mathcal{I}.
\end{equation}
The strategy profile $\alpha_*$ is called pure Nash equilibrium of the game. 
\noindent
\begin{definition} \textbf{Nash Equilibrium:} Given a game $\mathcal{G}$, a strategy profile $\alpha_*=(\alpha_*^i,\alpha_*^{-i})$ is a pure Nash equilibrium of $\mathcal{G}$ if and only if $
	u^i (\alpha_*^i,\alpha_*^{-i}) \geq u^i (\alpha^i,\alpha_*^{-i}),\allowbreak~~\forall \alpha^i \in \mathcal{A}^i,~~\forall i \in \mathcal{I}.$
\end{definition}
At a Nash equilibrium no player has a motivation to unilaterally deviate from its current state (\cite{nash}). 
\noindent 
%\begin{remark} Note that the potential function $ \Phi $ is player-independent, i.e., global. Thus, in a potential game, the set of pure Nash equilibriums can be found by identifying the local optima of the potential function.
%\end{remark}

A mixed strategy for player $i$ is defined when player $i$ randomly chooses between its actions in $\mathcal{A}^i$. Let $ x^i_{\alpha^i} $ be the probability that player $i$ selects action $ \alpha^i \in \mathcal{A}^i$ (the discrete version is denoted by $ X^i_{\alpha^i}$). Hence, player $i$'s mixed strategy is $x^i \in \mathcal{X}^i$ where $x^i=(x^i_{\alpha^i_1},...,x^i_{\alpha^i_{|\mathcal{A}^i|}})$ and $\mathcal{X}^i$ is a unit $ |\mathcal{A}^i| $-dimensional simplex $ \mathcal{X}^i=\{x \in \mathds{R}^{|\mathcal{A}^i|} ~s.t.~ \sum_{\alpha \in \mathcal{A}^i}~x_\alpha=1, ~x_\alpha \geq 0 \} $\nomenclature{$ \mathcal{X}^i $}{player $ i $'s mixed strategy space}. Likewise, we denote the mixed-strategy profile of all players by $x=(x^1,...,x^N) \in \mathcal{X}$ where the mixed strategy space is denoted by $ \mathcal{X}=\bigtimes_i \mathcal{X}^i$. 
\nomenclature{$ \mathcal{A}^i $}{player $ i $'s pure strategy space}
\nomenclature{$ \alpha^i $}{player $ i $'s pure action}
\nomenclature{$ x_\alpha^i $}{player $ i $'s mixed strategy probability corresponding to the pure action $ \alpha^i$}
\nomenclature{$ \mathcal{I} $}{set of all players}

A mixed-strategy Nash equilibrium is an $N$-tuple such that each player's mixed strategy maximizes its expected payoff if the strategies of the others are held fixed. Thus, each player's strategy is optimal against his opponents'. Let the expected utility of player $i$ be given as
\begin{equation}
\label{nash2}
u^i(x)=\sum_{\alpha \in \mathcal{A}} (\prod_{s \in \mathcal{I}} x^s_{\alpha^s}) u^i(\alpha^i,\alpha^{-i}),
\end{equation}
Then a mixed strategy profile $ x_* \in \mathcal{X} $ is a mixed strategy Nash equilibrium if for all players 
$
u^i(x^i_*,x^{-i}_*) \geq u^i(x^i,x^{-i}_*),~~\forall x^i \in \mathcal{X}^i,~~\forall i \in \mathcal{I}.
$
Such a Nash equilibrium is a fixed-point of the mixed-strategy best-response, or in other words, all players in a Nash equilibrium play their best response
$
x^i_* \in BR(x_*^{-i}),~~\forall i \in \mathcal{I}
$
where $BR(x^{-i})=\{x^i_* \in \mathcal{X}^i | u^i(x^i_*,x^{-i}) \geq u^i(x^i,x^{-i}),~\forall x^i \in \mathcal{X}^i\},$ is the best response set (\cite{newman}).

\subsection{Learning in Games}
\label{LiG}
Learning in games tries to relax assumptions of classical game theory on players' initial knowledge and belief about the game. In a game with learning, instead of immediately playing the perfect action, players adapt their strategies based on the outcomes of their past actions. In the following, we review two classes of learning in games: (1) log-linear learning and (2) reinforcement learning.

\subsubsection{Log-Linear Learning} 
In Log-Linear Learning (LLL), at each time step, only ``one" random player, e.g. player $i$, is allowed to alter its action. According to its mixed strategy $x^i$, player $i$ chooses a trial action from its ``entire" action set $\mathcal{A}^i$. In LLL, player $ i $'s mixed strategy or probability of action $\beta \in \mathcal{A}^i$ is updated by a Smooth Best Response (SBR) on $u^i$:
\vspace{-2mm}
\begin{equation}
\label{eq:12.8.1}
x_\beta^i=\dfrac{\exp(1/\tau~u^i(\beta,\alpha^{-i})}{\sum_{\gamma \in \mathcal{A}^i} \exp(1/\tau~u^i(\gamma,\alpha^{-i})}.
\end{equation}
where $ \tau $ is often called the temperature parameter that controls the smoothness of the SBR. The greater the temperature, the closer $ x^i $ is to the uniform distribution over player $ i $'s action space.  

Note that, each player $ i $ in LLL needs to know the utility of all actions in $ \mathcal{A}^i $, including those that are not played yet, and further actions of other players $ \alpha^{-i} $. With these assumptions, LLL can be modeled as a perturbed Markov process where the unperturbed Markov process is a best reply process. %The theory of resistance trees, which is reviewed in the following, is used in (\cite{marden}) to analyze the convergence of the Markov process associated with LLL. 
\noindent
\begin{definition} \textbf{Stochastically Stable State:} Let $Pr_\beta^\epsilon$ be the frequency or the probability with
	which action $\beta$ is played in the associated perturbed Markov process, where $\epsilon>0$ is the perturbation index. Action $\beta$ is then a stochastically stable state if:
	$
	\lim_{\epsilon \rightarrow 0} Pr_\beta^\epsilon= Pr_\beta
	$
	where $Pr_\beta$ is the corresponding probability in the unperturbed Markov process (\cite{young}).
\end{definition}
\emph{Synchronous Learning:(relaxing asynchrony assumption)} \\*
One of the basic assumptions in standard LLL is that only one random player is allowed to alter its action at each step. In (partial-) synchronous log-linear learning (SLLL) a group of players $ G \subset \mathcal{I} $ is selected to update its action based on the probability distribution $ rp \subset \Delta (2^\mathcal{I}) $; $ rp^G $ is defined as the probability that group $ G $ will be chosen and $ rp^i $ is defined as the probability that player $ i \in \mathcal{I} $ updates its action. The set of all groups with $ rp^G>0 $ is denoted by $ \bar{G} $. In an independent revision process, each player independently decides whether to revise his strategy by LLL rule. SLLL is proved to converge under certain assumptions (\cite{marden}).
\vspace{1mm}
\noindent\emph{Constrained Action Set:(relaxing completeness assumption)} \\* 
Standard LLL requires each player $i$ to have access to all available actions in $\mathcal{A}^i$. In the case when player $i$ has no free access to every action in $\mathcal{A}^i$, its action set is ``constrained" and is denoted by $\mathcal{A}^i_c:\mathcal{A}^i\rightarrow 2^{\mathcal{A}^i}$. With a constrained action set, players may be trapped in local sub-optimal equilibria since the entire $\mathcal{A}^i$ is not available to player $i$ at each move. Thus, stochastically stable states may not be potential maximizers. Binary log-linear learning (BLLL) is a variant of standard LLL which provides a solution to this issue.

%In a BLLL scheme, at each time $n$, only one random player $ i $ is allowed to alter his action while all the other players must repeat their current actions. Player $ i $ selects one trial action $ \alpha_T^i $ uniformly randomly from his constrained action set $ \mathcal{A}_c^i(\alpha^i(n)) $. Player $ i $'s mixed strategy is then updated by
%\begin{align}
%\label{eq:12.8.4}
%\begin{aligned}
%{}&X^{i}_{\alpha^i(n)}(n)= \dfrac{\exp(\dfrac{1}{\tau} u^i(\alpha^i(n),\alpha^{-i}(n)))}{\exp(\dfrac{1}{\tau} u^i(\alpha^i(n),\alpha^{-i}(n)))+\exp(\dfrac{1}{\tau} u^i (\alpha_T^i,\alpha^{-i}(n)))},
%\end{aligned}
%\end{align}
%\begin{align}
%\label{eq:12.8.5}
%\begin{aligned}
%{}&~~X^{i}_{\alpha_T^i}(n)=\dfrac{\exp(\dfrac{1}{\tau} u^i (\alpha_T^i,\alpha^{-i}(n)))}{\exp(\dfrac{1}{\tau} u^i(\alpha^i(n-1),\alpha^{-i}(n)))+\exp(\dfrac{1}{\tau} u^i (\alpha_T^i,\alpha^{-i}(n)))},
%\end{aligned}
%\end{align}
%where $ X^{i}_{\alpha^i(n)} $ is the probability that player $ i $ remains on his current action $\alpha^i(n)$ and $X^{i}_{\alpha_T^i}$ is the probability that player $ i $ selects the trial action $ \alpha_T^i $. 
\begin{assumption} \label{ass1}For each player $ i $ and for any action pair $ \alpha^i(1),\alpha^i(m) \in \mathcal{A}^i $ there exists a sequence of actions $ \alpha^i(1)~\rightarrow~\alpha^i(2)~\rightarrow~...~\rightarrow~\alpha^i(m) $ satisfying $ \alpha^i(k) \in \mathcal{A}_c^i(\alpha^i({k-1})),~\forall k \in \{2,...,m\}.$\end{assumption}
\begin{assumption} \label{ass2}For each player $ i $ and for any action pair $ \alpha^i(1),\alpha^i(2) \in \mathcal{A}^i $, $ \alpha^i(2) \in \mathcal{A}_c^i(\alpha^i(1))~\allowbreak\Leftrightarrow~ \alpha^i(1) \in \mathcal{A}_c^i(\alpha^i(2)).$\end{assumption}
The following theorem studies the convergence of BLLL in potential games under Assumptions \ref{ass1} and \ref{ass2}. 
\begin{theorem}In a finite $N$-player potential game satisfying Assumptions \ref{ass1} and \ref{ass2} and with potential function $\Phi: \mathcal{A} \mapsto R$, if all players adhere to BLLL, then the stochastically stable states are the set of potential maximizers (\cite{marden}).
\end{theorem}

\subsubsection{Reinforcement Learning}
Reinforcement Learning (RL) is another variant of learning algorithms that we consider in this paper. RL discusses how to map actions' reward to players' action so that the accumulated reward is maximized. Players are not told which actions to take but instead they have to discover which actions yield the highest reward by ``exploring" the environment (\cite{sutton}). RL only requires players to observe their own ongoing payoffs, so they do not need to monitor their opponents' strategies or predict payoffs of actions that they did not play. In the following we present a technical background on RL and its application in MASs.

\vspace{1mm}
\noindent\emph{Aggregation}: \\*
In RL each player uses a score variable, as a memory, to store and track past events. We denote player $ i $'s score vector by $ p^i \in \mathcal{P}^i$ where $ \mathcal{P}^i $ is player $ i $'s score space. It is common to assume that the game rewards can be stochastic, i.e., an action profile does not always result in the same deterministic utility. Therefore, actions need to be sampled, i.e. aggregated, repeatedly or continuously.  A common form of continuous aggregation rule is the exponential discounted model:
\begin{equation}
\label{eq:1.n}
\begin{split}
p_{\beta}^i(t)=p_{\beta}^i(0) \lambda^t+\int_0^t \lambda^{t-s} u^i(\beta,\alpha^{-i}) ds,~ (\beta,\alpha^{-i}) \in \mathcal{A},
\end{split}
\end{equation}
where $ p_{\beta}^i $ is action $ \beta $'s aggregated score and $\lambda>0$ \nomenclature{$\lambda$}{discount rate}is the model's discount rate. $\lambda$ can be alternatively defined via $ T=log(1/\lambda) $. 
%If $p_{\beta}^i(0)=0$ then
%\begin{itemize}
%	\item For $ \lambda \in (0,1) $, the model in (\ref{eq:1.n}) allocates more weight (exponentially) to recent observations ($ T>0 $),
%	\item If $ \lambda=1 $, we have a uniform aggregation, i.e., all past observations are treated uniformly ($ T=0 $),
%	\item For $ \lambda > 1 $, the model considers more weight for older observations ($ T<0 $).
%\end{itemize}
The choice of $ T $ affects the learning dynamics and the process of finding the estimated Nash equilibrium. The discount rate has a double role in RL: (1) It determines the weight that players give to their past observations. (2) $ T $ reflects the rationality of the players in choosing their actions and consequently the accuracy of players' stationary points in being the true Nash equilibrium. Additionally, discounting implies that the score variable $ p_{\beta}^i(t) $ will remain bounded which consequently prevent the agents' mixed strategies from approaching the boundaries $ \mathcal{X} $ (\cite{num1}). By differentiating (\ref{eq:1.n}), and assuming $p_{\beta}^i(0)=0$, we obtain the following score dynamics
\begin{equation}
\label{eq:2.n}
\begin{split}
{\dot{p}}_\beta^i=u^i(\beta,\alpha^{-i})-T p_\beta^i.
\end{split}
\end{equation}
By applying the first-order Euler discretization on (\ref{eq:2.n}) we obtain:
\begin{align}
\label{eq:12.n}
\begin{aligned}
{}& P_\beta^i (n+1)=P_\beta^i(n)+\mu(n) [u^i (\beta,\alpha^ {-i})-T P_\beta^i(n)],
\end{aligned}
\end{align}
where $ n $ is the iteration number, $ \mu(n) $\nomenclature{$ \mu` $}{discretization step size} is the discretization step size and $ P_\beta^i(n) $ is the discrete equivalent of $ p_\beta^i(t) $. A stochastic approximation of the discrete dynamics requires diminishing step sizes such that $ \sum_n \mu(n)=\infty $ and $ \sum_n {\mu(n)}^2<\infty $ (\cite{benaim}).\\*
\\*
\emph{Choice Map}: \\*
In the action selection step players decide how to exploit the score variable to choose a strategy against the environment, e.g. according to:
%One natural method is to select a strategy that maximizes the expected score via the mapping
%\begin{equation}
%\label{eq:5}
%\begin{split}
%BR(p^i)=\arg\!\max_{x^i \in \mathcal{X}^i} \sum_{\beta \in \mathcal{A}^i} x_{\beta}^i p_\beta^i,
%\end{split}
%\end{equation}
%where $ p^i $ is the player $ i $'s score vector. However, this simple BR map may raise several problems: (1) Equation (\ref{eq:5}) can become a multi-valued mapping if at least two score variables $ p_\alpha^i $, $ p_\beta^i \in \mathcal{P}^i$ happen to be equal. (2) A simple BR generally forces RL to converge to pure strategies while we may be interested in non-pure equilibria. (3) Such a choice map may lead to a discontinuous trajectories of play due to the sudden jumps in the BR correspondence.
%To overcome these issues, a smooth strongly convex penalty function ($ h^i: \mathcal{X}^i \mapsto R$) is usually added to (\ref{eq:5}) to regularize the choice map:
\begin{equation}
\label{eq:6}
\begin{split}
SBR(p^i)= \arg\!\max_{x^i \in \mathcal{X}^i} \sum_{\beta \in \mathcal{A}^i} [x_\beta^i p_\beta^i-h^i(x^i)].
\end{split}
\end{equation}
This choice\nomenclature{$ SBR(p) $}{smoothed best response map} model is often called ``Smoothed Best Response (SBR) map" or ``quantal response function" where the penalty function $ h^i $ in (\ref{eq:6}) has to have the following properties:
\begin{enumerate}
	\item $ h^i $ is finite except on the relative boundaries of $ \mathcal{X}^i $,
	\item $ h^i $ is continuous on $\mathcal{X}^i  $, smooth on relative interior of $ \mathcal{X}^i $ and $ |dh^i(x^i)| \rightarrow +\infty $ as $ x^i $ approaches to the boundaries of $ \mathcal{X}^i $,
	\item $ h^i $ is convex on $ \mathcal{X}^i $ and strongly convex on\nomenclature{$ h $}{penalty function} relative interior of $ \mathcal{X}^i $.
\end{enumerate} 
%Some of the famous penalty functions are:
%\begin{enumerate}
%\item The Gibbs entropy:$~~~~~~~h(x)=\sum_\beta x_\beta \log x_\beta $,
%\item The Tsallis entropy:$~~~~~~h(x)=(1-q)^{-1} \sum_\beta(x_\beta-x_\beta^q),~~0<q \leq 1 $,
%\item The Burg entropy:$~~~~~~~~h(x)=-\sum_\beta x \log x_\beta $.
%\end{enumerate}
The choice map (\ref{eq:6}) actually discourages player $ i $ from choosing an action from boundaries of $ \mathcal{X}^i $, i.e. from choosing pure strategies. The most prominent SBR map is the ``logit" map based on using Gibbs entropy in (\ref{eq:6}),
\begin{equation}
\label{eq:6.5}
\begin{split}
x_\alpha^i=\Big[SBR(p^i)\Big]_\alpha=\dfrac{\exp(p_\alpha^i)}{\sum_{\beta \in \mathcal{A}^i}~\exp(p_\beta^i)}.
\end{split}
\end{equation}
It is not always easy to write a closed form of the choice map and the Gibbs entropy is an exception (\cite{num1}).\\
\noindent
In order to discretize (\ref{eq:6}) we again apply first-order Euler discretization:
\begin{align}
\label{eq:12.n.2}
\begin{aligned}
X^i(n+1)=SBR(P^i(n)),
\end{aligned}
\end{align}
where $ X^i $ is the discrete equivalent of $ x^i $.

At this point all the necessary background is presented and in the following we are going to discuss our proposed learning algorithms.

\section{Partial-Synchronous Binary Log-Linear Learning}
\label{sblll}
In this section, we present a modified LLL algorithm in which both assumptions on asynchrony and complete action set are relaxed. This means that in a Partial-Synchronous Binary Log-Linear Learning (P-SBLLL) scheme agents can learn simultaneously while their available action sets are constrained. This simultaneous learning presumably increases the BLLL learning rate in the MAS problem.

In P-SBLLL algorithm, we propose that at each time $ n $, a set of players $ S(n) \subseteq \mathcal{I} $ independently update their actions according to each player $ i $'s revision probability $ rp^i:\mathcal{A}^i \rightarrow (0,1) $. The revision probability $rp^i$ is the probability with which agent $i$ wakes up to update its action. All the other players $ \mathcal{I} \setminus S(n) $ must repeat their current actions. \nomenclature{$ rp^i $}{player $i$'s revision probability}

Each player $ i \in S(n) $ selects one trial action $ \alpha_T^i $ uniformly randomly from its constrained action set $ \mathcal{A}_c^i(\alpha^i(n)) $. Then, player $ i $'s mixed strategy is 
\begin{align}
\label{eq:12.8.4n}
\begin{aligned}
{}&X^{i}_{\alpha^i(n)}(n)= \dfrac{\exp(\dfrac{1}{\tau} u^i(\alpha(n)))}{\exp(\dfrac{1}{\tau} u^i(\alpha(n)))+\exp(\dfrac{1}{\tau} u^i (\alpha_T))},
\end{aligned}
\end{align}
\begin{align}
\label{eq:12.8.5n}
\begin{aligned}
{}&X^{i}_{\alpha_T^i}(n)=\dfrac{\exp(\dfrac{1}{\tau} u^i (\alpha_T))}{\exp(\dfrac{1}{\tau} u^i(\alpha(n)))+\exp(\dfrac{1}{\tau} u^i (\alpha_T))},
\end{aligned}
\end{align}
where $ \alpha_T$ is the action profile for which each player  $ i \in S(n) $ updates its action to $\alpha_T^i$ and all the other players $ \mathcal{I} \setminus S(n) $ repeat their actions. In the following, we analyze the convergence of the proposed algorithm. 

\subsection{P-SBLLL's Convergence Analysis}
From the theory of resistance trees, we use the useful relationship between stochastically stable states and potential maximizing states. We make the following assumption:

\begin{assumption} \label{ass3} For each player $ i $, and for each action $ \alpha^i $, the revision probability $ rp^i $ must be bounded and $ 0<rp^i(\alpha^i(n))<1 $.\end{assumption}
\begin{lemma}
	\label{lemnr1} 
	Under Assumption \ref{ass3}, P-SBLLL induces a perturbed Markov process where the resistance of any feasible transition $ \alpha_1 \in \mathcal{A} \rightarrow \alpha_2 \in \mathcal{A}$ with deviating set of players $S$ is
	\begin{equation}
	\label{eq:12.8.60}
	\begin{split}
	R(\alpha_1 \rightarrow \alpha_2)= \sum_{i \in S} \max\{u^i(\alpha_1),u^i(\alpha_2)\}-u^i(\alpha_2),
	\end{split}
	\end{equation}
	where each deviating player $ i\in S $ selects its action based on $ \mathcal{A}^i_c $.
\end{lemma}
\textbf{Proof:} Let $ P_\epsilon,~\epsilon>0$ denote the perturbed transition matrix. The probability of transition from $ \alpha_1 $ to $ \alpha_2 $ is
\begin{equation}
\label{eq:12.8.61}
\begin{split}
P_{\epsilon(\alpha_1 \rightarrow \alpha_2)}=\prod_{i \in S}~\dfrac{rp^i(\alpha_1^i)}{|\mathcal{A}_c^i(\alpha_1^i)|}~\prod_{j \in \mathcal{I}\setminus S} (1-rp^j(\alpha_1^j))~\prod_{i \in S} \dfrac{\epsilon^{-u^i(\alpha_2)}}{\epsilon^{-u^i(\alpha_1)}+\epsilon^{-u^i(\alpha_2)}},
\end{split}
\end{equation}
where $ \epsilon:=e^{-1/\tau} $. The first term $\prod_{i \in S}~\dfrac{rp^i(\alpha_1^i)}{|\mathcal{A}_c^i(\alpha_1^i)|}$ represents the probability that all the players in $ S $ wake up to change their actions from $ \alpha_1 $ to $ \alpha_2 $. The second term $\prod_{j \in \mathcal{I}\setminus S} (1-rp^j(\alpha_1^j))$ is the probability that the players in $ \mathcal{I} \setminus S $ stay asleep. The last term $\prod_{i \in S} \dfrac{\epsilon^{-u^i(\alpha_2)}}{\epsilon^{-u^i(\alpha_1)}+\epsilon^{-u^i(\alpha_2)}}$ is the binary SBR over $\alpha_1$ and $\alpha_2$. Next define the maximum utility of player $ i $ for any two action profiles $ \alpha_1 $ and $ \alpha_2 $ as 
$
V^i(\alpha_1,\alpha_2)=\max\{u^i(\alpha_1),u^i(\alpha_2)\}.
$
By multiplying the numerator and denominator of (\ref{eq:12.8.61}) by $ \prod_{i \in S}~ \epsilon^{V^i(\alpha_1,\alpha_2)} $ we obtain
\begin{equation}
\label{eq:12.8.62}
\begin{split}
{}&P_{\epsilon(\alpha_1 \rightarrow \alpha_2)}=\\
&\prod_{i \in S}~\dfrac{rp^i(\alpha_1^i)}{|\mathcal{A}_c^i(\alpha_1^i)|}~\prod_{j \in \mathcal{I}\setminus S} (1-rp^j(\alpha_1^j))~\prod_{i \in S} \dfrac{\epsilon^{V^i(\alpha_1,\alpha_2)-u^i(\alpha_2)}}{\epsilon^{V^i(\alpha_1,\alpha_2)-u^i(\alpha_1)}+\epsilon^{V^i(\alpha_1,\alpha_2)-u^i(\alpha_2)}}.
\end{split}
\end{equation}
Dividing (\ref{eq:12.8.62}) by $ \epsilon^{\sum_{i \in S} V^i(\alpha_1,\alpha_2)-u^i(\alpha_2)} $ yields$ \dfrac{P_{\epsilon(\alpha_1 \rightarrow \alpha_2)}}{\epsilon^{\sum_{i \in S} V^i(\alpha_1,\alpha_2)-u^i(\alpha_2)}} $ to be
\begin{equation}
\label{eq:12.8.620}
\begin{split}
\prod_{i \in S}\dfrac{rp^i(\alpha_1^i)}{|\mathcal{A}_c^i(\alpha_1^i)|}\prod_{j \in \mathcal{I}\setminus S} (1-rp^j(\alpha_1^j))\prod_{i \in S} \dfrac{1}{\epsilon^{V^i(\alpha_1,\alpha_2)-u^i(\alpha_1)}+\epsilon^{V^i(\alpha_1,\alpha_2)-u^i(\alpha_2)}}.
\end{split}
\end{equation}
According to theory of resistance trees (\cite{young}), if $$0< \lim_{\epsilon \rightarrow 0}\allowbreak~\dfrac{P_{\epsilon(\alpha_1 \rightarrow \alpha_2)}}{\epsilon^{\sum_{i \in S} V^i(\alpha_1,\alpha_2)-u^i(\alpha_2)}}\allowbreak < \infty $$ then our claim about $R(\alpha_1 \rightarrow \alpha_2)$ in (\ref{eq:12.8.60}) is true. Considering the definition of $ V^i(\alpha_1,\alpha_2) $, we know that for each player $i$, either $V^i(\alpha_1,\alpha_2)-u^i(\alpha_1)$ or $V^i(\alpha_1,\alpha_2)-u^i(\alpha_2)$ is zero and the other one is a positive real number. Thus, as $\epsilon \rightarrow 0$, $\prod_{i \in S} \dfrac{1}{\epsilon^{V^i(\alpha_1,\alpha_2)-u^i(\alpha_1)}+\epsilon^{V^i(\alpha_1,\alpha_2)-u^i(\alpha_2)}}$ in (\ref{eq:12.8.620}) approaches $1$ and 
\begin{equation}
\label{eq:12.8.63}
\begin{split}
\lim_{\epsilon \rightarrow 0}~\dfrac{P_{\epsilon(\alpha_1 \rightarrow \alpha_2)}}{\epsilon^{\sum_{i \in S} V^i(\alpha_1,\alpha_2)-u^i(\alpha_2)}}=\prod_{i \in S}~\dfrac{rp^i(\alpha_1^i)}{|\mathcal{A}_c^i(\alpha_1^i)|}~\prod_{j \in \mathcal{I}\setminus S} (1-rp^j(\alpha_1^j)).
\end{split}
\end{equation}
From Assumption \ref{ass3},$~ \prod_{i \in S}~\dfrac{rp^i(\alpha_1^i)}{|\mathcal{A}_c^i(\alpha_1^i)|} $ and $ \prod_{j \in \mathcal{I}\setminus S} (1-rp^j(\alpha_1^j)) $ are finite positive real numbers. Hence,
$
0 < \lim_{\epsilon \rightarrow 0}~\dfrac{P_{\epsilon(\alpha_1 \rightarrow \alpha_2)}}{\epsilon^{\sum_{i \in S} max\{u^i(\alpha_1),u^i(\alpha_2)\}-u^i(\alpha_2)}} < \infty.
$
Therefore, the process is a perturbed Markov process where the resistance of the transition $ \alpha_1 $ to $ \alpha_2 $ is 
$
R(\alpha_1 \rightarrow \alpha_2)= \sum_{i \in S} \max\{u^i(\alpha_1),u^i(\alpha_2)\}-u^i(\alpha_2).
$
$ ~\hfill\square $
\begin{definition} \textbf{Separable Utility Function:} Player $i$'s utility function $u^i$ is called separable if $u^i$ only depends on player $i$'s action $\alpha^i$. 
\end{definition}
\begin{lemma} 
	\label{lemnr2}
	Consider any finite $ N $-player potential game where all the players adhere to P-SBLLL and the potential function is defined as $ \Phi: \mathcal{A} \rightarrow \mathds{R} $. Assume that players' utility functions are separable. For any feasible transition $ \alpha_1 \in \mathcal{A} \rightarrow \alpha_2 \in \mathcal{A}$ with deviating set of players $ S $, the following holds:
	\begin{equation}
	\label{eq:12.8.65}
	\begin{split}
	R(\alpha_1 \rightarrow \alpha_2) - R(\alpha_2 \rightarrow \alpha_1)= \Phi(\alpha_1)-\Phi(\alpha_2).
	\end{split}
	\end{equation}
\end{lemma}
\textbf{Proof:} From Lemma \ref{lemnr1},
$
R(\alpha_1 \rightarrow \alpha_2)= \sum_{i \in S_{12}} \max\{u^i(\alpha_1),u^i(\alpha_2)\}-u^i(\alpha_2), 
$
and
$
R(\alpha_2 \rightarrow \alpha_1)= \sum_{i \in S_{21}} \max\{u^i(\alpha_2),u^i(\alpha_1)\}-u^i(\alpha_1),
$
where $ S_{12} $ is the set of deviating players during the transition $ \alpha_1 \rightarrow \alpha_2 $ and $ S_{21} $ is the set of deviating players in the transition $ \alpha_2 \rightarrow \alpha_1 $. By Assumption \ref{ass2}, if the transition $ \alpha_1 \rightarrow \alpha_2 $ is possible then there exist a reverse transition $ \alpha_2 \rightarrow \alpha_1 $. Clearly, the same set of deviating players is needed for both $ \alpha_1 \rightarrow \alpha_2 $ and $ \alpha_2 \rightarrow \alpha_1 $, i.e. $ S_{12} = S_{21}=S$. Therefore:
%$$
%\alpha_1 \rightarrow \alpha_{2} = \{\alpha_1=\alpha_{k_{0}} \rightarrow \alpha_{k_1} \rightarrow \alpha_{k_2} \rightarrow ... \rightarrow \alpha_{k_{q-1}} \rightarrow \alpha_{k_q}=\alpha_{2} \}
%$$
%where $q=|S|$. Thus:
%$$
%R(\alpha_1 \rightarrow \alpha_2)= max\{u^{s_1}(\alpha_{k_{0}}),u^{s_1}(\alpha_{k_1})\}-u^{s_1}(\alpha_{k_1})+...+max\{u^{s_q}(\alpha_{k_{q-1}}),u^{s_q}(\alpha_{k_q})\}-u^{s_q}(\alpha_{k_q})
%$$
%For the reverse transition $ \alpha_2 \rightarrow \alpha_1 $ we have:
%$$
%R(\alpha_2 \rightarrow \alpha_1)= max\{u^{s_q}(\alpha_{k_{q-1}}),u^{s_q}(\alpha_{k_q})\}-u^{s_q}(\alpha_{k_{q-1}})+...+max\{u^{s_1}(\alpha_{k_{0}}),u^{s_1}(\alpha_{k_1})\}-u^{s_1}(\alpha_{k_0})
%$$
%and
\begin{align}
\label{eq:12.8.66}
\begin{aligned}
{}&R(\alpha_1 \rightarrow \alpha_2) - R(\alpha_2 \rightarrow \alpha_1)=\\
& \Big[\sum_{i \in S_{12}} \max\{u^i(\alpha_1),u^i(\alpha_2)\}-u^i(\alpha_2)\Big]-\Big[\sum_{i \in S_{21}} \max\{u^i(\alpha_2),u^i(\alpha_1)\}-u^i(\alpha_1)\Big]=\\
&\sum_{i \in S} \Big[ \max\{u^i(\alpha_1),u^i(\alpha_2)\}-u^i(\alpha_2)  - \max\{u^i(\alpha_2),u^i(\alpha_1)\}+u^i(\alpha_1)\Big].
\end{aligned}
\end{align}
By canceling identical terms $max\{u^i(\alpha_2),u^i(\alpha_1)\}$,
$
R(\alpha_1 \rightarrow \alpha_2) - R(\alpha_2 \rightarrow \alpha_1)= \sum_{i \in S} u^i(\alpha_1)-u^i(\alpha_2).
$
Since players' utility functions are separable, for any player $i$, $u^i(\alpha_1)-u^i(\alpha_2)=u^i(\alpha_1^i,\alpha_1^{-i})-u^i(\alpha_2^i,\alpha_1^{-i})$. From Definition \ref{PG} we have 
$
u^i(\alpha_1^i,\alpha_1^{-i})-u^i(\alpha_2^i,\alpha_1^{-i})=\Phi(\alpha_1^i,\alpha_1^{-i})-\Phi(\alpha_2^i,\alpha_1^{-i})
$
and it is easy to show that
$
\sum_{i \in S} u^i(\alpha_1)-u^i(\alpha_2)=\Phi(\alpha_1)-\Phi(\alpha_2).
$
%Now divide $\alpha_1 \rightarrow \alpha_2$ into a sequence of sub-edges in which only one agent is deviating. Assuming $ S=\{i,i+1,i+2,...,i+|S|\} $ as the set of deviating players for this transition, the expanded from of this edge is
%$$
%\alpha_1 \rightarrow \alpha_{2} = \{\alpha_1=\alpha_{k_{0}} \rightarrow (\alpha_2^i,\alpha_1^{-i})=\alpha_{k_1} \rightarrow \alpha_{k_2} \rightarrow ... \rightarrow \alpha_{k_{q-1}} \rightarrow \alpha_{k_q}=\alpha_{2} \},
%$$
%where $ \alpha_{k_{j}} \rightarrow \alpha_{k_{j+1}} $ is a sub-edge in which only player $i+j$th is deviating. Thus, $R(\alpha_1 \rightarrow \alpha_2) - R(\alpha_2 \rightarrow \alpha_1)$ in (\ref{eq:12.8.66.12}) can be written as:
%\begin{align}
%\label{eq:12.8.66.12.p.55}
%\begin{aligned}
%{}&R(\alpha_1 \rightarrow \alpha_2) - R(\alpha_2 \rightarrow \alpha_1)= \sum_{i \in S} \mathbold{u^i(\alpha_1)}-u^i(\alpha_{k_1})+u^i(\alpha_{k_1})-
%u^{i+1}(\alpha_{k_2})+u^{i+1}(\alpha_{k_2})\\
%&-...+u^{i+|S|-1}(\alpha_{k_{q-1}})-\mathbold{u^i(\alpha_2)}.
%\end{aligned}
%\end{align}
%Since players' utility functions are local, 
%\begin{align}
%\label{eq:12.8.66.5}
%\begin{aligned}
%{}&R(\alpha_1 \rightarrow \alpha_2) - R(\alpha_2 \rightarrow \alpha_1)= \sum_{i \in S} u^i(\alpha_1)-u^i(\alpha_2)= \sum_{i \in S} \phi(\alpha_1)-\phi(\alpha_2).
%\end{aligned}
%\end{align}
%Since $\phi$ and action profiles $\alpha_1$ and $\alpha_2$ are not dependent on $i$
%\begin{align}
%\label{eq:12.8.66.55}
%\begin{aligned}
%{}&\sum_{i \in S} \phi(\alpha_1)-\phi(\alpha_2)=|S|(\phi(\alpha_1)-\phi(\alpha_2)).
%\end{aligned}
%\end{align}
$ ~\hfill\square $\\
Next, we prove that in separable games, the stable states of the algorithm maximize the potential function. In other words, the stable states are the optimal states. 
\begin{proposition}
	\label{prop1} 
	Consider any finite $ N $-player potential game satisfying Assumptions \ref{ass1}, \ref{ass2} and \ref{ass3}. Let the potential function be $ \Phi: \mathcal{A} \rightarrow \mathds{R} $ and assume that all players adhere to P-SBLLL. If the utility functions for all players are separable, then the stochastically stable states are the set of potential maximizers.
\end{proposition} 
\textbf{Proof:} We first show that for any path $ \Omega $, defined as a sequence of action profiles $ \Omega:= \{\alpha_0 \rightarrow \alpha_1 \rightarrow ... \rightarrow \alpha_m \} $ and its reverse path $ \Omega^R:= \{\alpha_m \rightarrow \alpha_{m-1} \rightarrow ... \rightarrow \alpha_0 \} $, the resistance difference is
$
R(\Omega)-R(\Omega^R)=\Phi(\alpha_0)-\Phi(\alpha_m).
$
where 
\begin{equation}
\label{123}
R(\Omega):=\sum_{k=0}^{m-1} R(\alpha_k \rightarrow \alpha_{k+1}),~R(\Omega^R):=\sum_{k=0}^{m-1} R(\alpha_{k+1} \rightarrow \alpha_{k}).
\end{equation}

\begin{figure}[!t]
	\centering
	\subfloat[Tree $T$ rooted at $\alpha$]{{\includegraphics[scale=0.15]{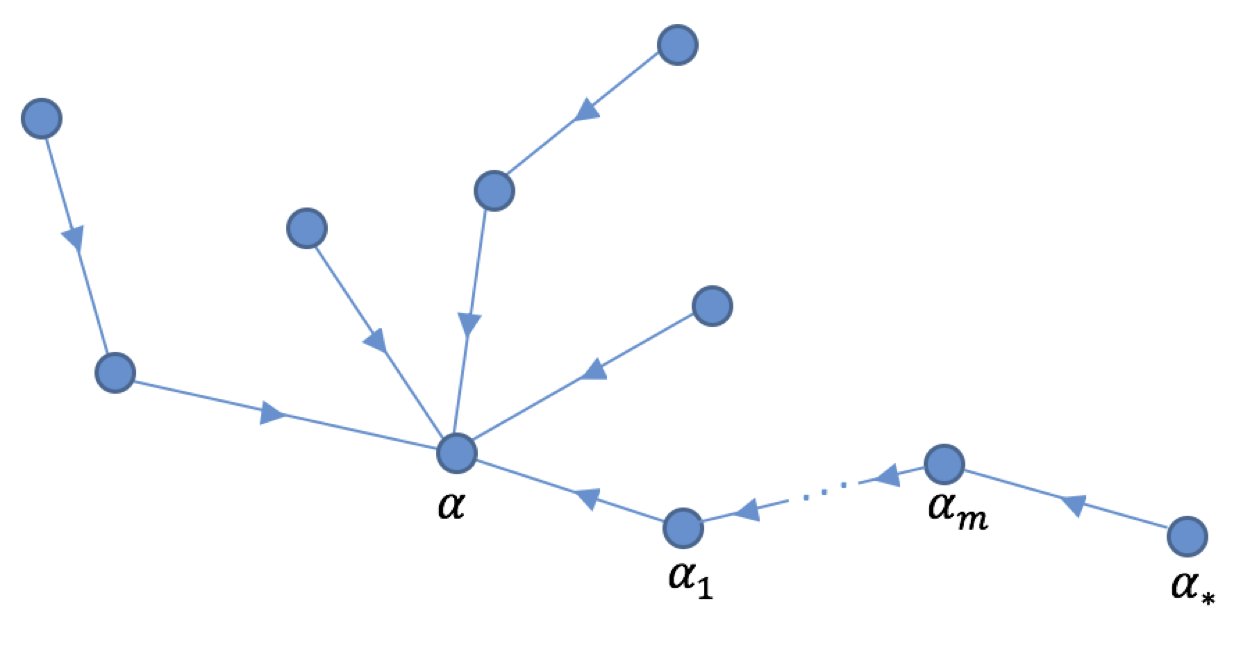}}}
	~
	\subfloat[$\Omega_p$ and its reverse path $\Omega_p^R$]{{\includegraphics[scale=0.15]{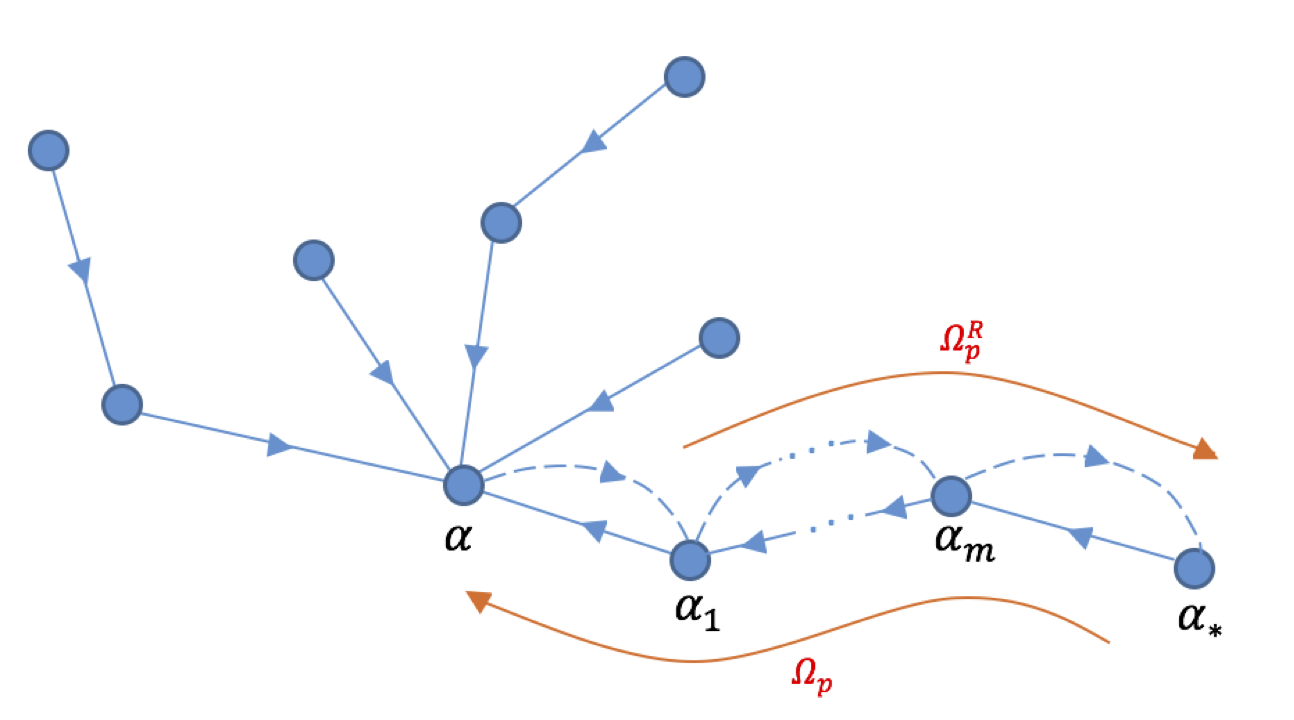}}}
	\\
	\subfloat[Tree $T'$ rooted at $\alpha_*$]{{\includegraphics[scale=0.15]{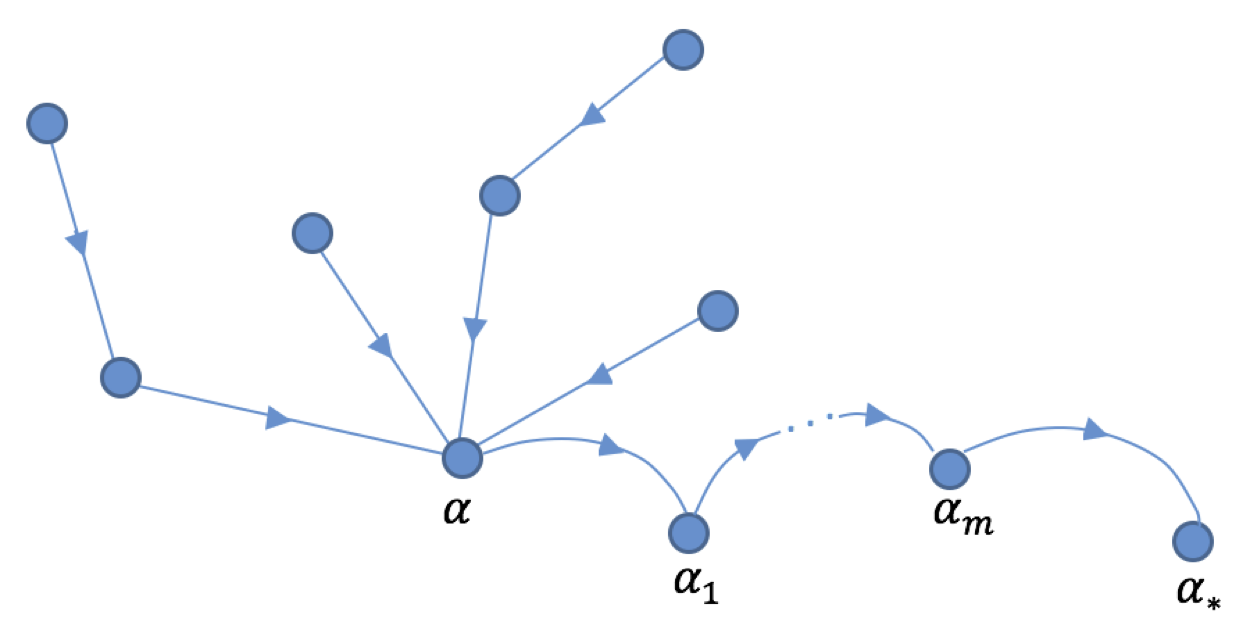}}}
	\caption{Resistance Trees}
	\label{treeN}
\end{figure}

\noindent Assuming $ S $ as the set of deviating players for the edge $ \alpha_k \rightarrow \alpha_{k+1} $ in $\Omega$, the expanded form of this edge can be written as
$
\alpha_k \rightarrow \alpha_{k+1} = \{\alpha_k=\alpha_{k_{0}} \rightarrow \alpha_{k_1} \rightarrow \alpha_{k_2} \rightarrow ... \rightarrow \alpha_{k_{q-1}} \rightarrow \alpha_{k_q}=\alpha_{k+1} \},
$
where $ \alpha_{k_i} \rightarrow \alpha_{k_{i+1}} $ is a sub-edge in which only one player is deviating and $ q=|S| $. From Lemma \ref{lemnr2} 
$
R(\alpha_{k_i} \rightarrow \alpha_{k_{i+1}}) - R(\alpha_{k_{i+1}} \rightarrow \alpha_{k_i})= \Phi(\alpha_{k_i})-\Phi(\alpha_{k_{i+1}}).
$
Note that $ |S|=1 $ for each sub-edge. Consequently for each edge $ \alpha_k \rightarrow \alpha_{k+1} = \{\alpha_k=\alpha_{k_{0}} \rightarrow \alpha_{k_1} \rightarrow \alpha_{k_2} \rightarrow ... \rightarrow \alpha_{k_{q-1}} \rightarrow \alpha_{k_q}=\alpha_{k+1} \} $ we obtain $$R(\alpha_k \allowbreak \rightarrow \alpha_{k+1}) - R(\alpha_{k+1} \rightarrow \alpha_k)=$$
$$
 \Phi(\alpha_{k_0})-\Phi(\alpha_{{k_1}})+\Phi(\alpha_{k_1})-\allowbreak\Phi(\alpha_{{k_2}})+...+\Phi(\alpha_{k_{q-1}})-\Phi(\alpha_{k_{q}}).
$$
By canceling the identical terms,
\begin{align}
\label{eq:12.8.67}
\begin{aligned}
R(\alpha_k \rightarrow \alpha_{k+1}) - R(\alpha_{k+1} \rightarrow \alpha_k)=\Phi(\alpha_{k_0})-\Phi(\alpha_{k_{q}})=
\Phi(\alpha_{k})-\Phi(\alpha_{{k+1}}).
\end{aligned}
\end{align}
Comparing (\ref{eq:12.8.67}) and (\ref{eq:12.8.65}) implies that if the utility functions of all players are separable, the number of deviating players does not affect the resistance change between forward and backward transitions. Finally, we sum up the resistance difference in (\ref{eq:12.8.67}) for all pairs $(\alpha_k,\alpha_{k+1}),~k \in \{0,1,...,m-1\}$:
$
\sum_{k=0}^{m-1} R(\alpha_k \rightarrow \alpha_{k+1}) - R(\alpha_{k+1} \rightarrow \alpha_k)=\sum_{k=0}^{m-1} \Phi(\alpha_{k})-\Phi(\alpha_{{k+1}}),
$
or equivalently
\begin{align}
\label{eq:12.8.67.6}
\begin{aligned}
\sum_{k=0}^{m-1} R(\alpha_k \rightarrow \alpha_{k+1}) - \sum_{k=0}^{m-1} R(\alpha_{k+1} \rightarrow \alpha_k)=\sum_{k=0}^{m-1} \Phi(\alpha_{k})-\Phi(\alpha_{{k+1}}).
\end{aligned}
\end{align}
From (\ref{123}) we have $\sum_{k=0}^{m-1} R(\alpha_k \rightarrow \alpha_{k+1})$ is the resistance over the path $\Omega$, i.e. $R(\Omega)$ and $\sum_{k=0}^{m-1} R(\alpha_{k+1} \rightarrow \alpha_k)$ is the resistance over the reverse path $\Omega^R$, i.e. $R(\Omega^R)$. Furthermore, it is easy to show $\sum_{k=0}^{m-1} \Phi(\alpha_{k})-\Phi(\alpha_{{k+1}})=
\Phi(\alpha_0)-\Phi(\alpha_m)$. Consequently,
\begin{align}
\label{eq:12.8.68}
\begin{aligned}
R(\Omega)-R(\Omega^R)=\Phi(\alpha_0)-\Phi(\alpha_m).
\end{aligned}
\end{align}
Now assume that an action profile $ \alpha $ is a stochastically stable state. Therefore, there exist a tree $ T $ rooted at $\alpha$ (Fig. \ref{treeN}.a) for which the resistance is minimum among the trees rooted at other states. We use contradiction to prove our claim. As the contradiction assumption, suppose the action profile $ \alpha $ does not maximize the potential and let $ \alpha_* $ be the action profile that maximizes the potential function. Since $ T $ is rooted at $ \alpha $ and from Assumption \ref{ass1}, there exist a path $ \Omega_p $ from $ \alpha_* $ to $ \alpha $ as
$ 
\Omega_p = \{\alpha_* \rightarrow \alpha_1 \rightarrow ... \rightarrow \alpha \}.
$
Consider the reverse path $ \Omega_p^R  $ from $ \alpha $ to $ \alpha_* $ (Fig. \ref{treeN}.b)
$ 
\Omega_p^R = \{\alpha \rightarrow \alpha_m \rightarrow ... \rightarrow \alpha_* \}.
$ 
We can construct a new tree $ T' $ rooted at $ \alpha_* $ by adding the edges of $  \Omega_p^R  $ to $ T $ and removing the edges of $  \Omega_p $ (Fig. \ref{treeN}.c). The resistance of the new tree $T'$ is
$
R(T')=R(T)+R( \Omega_p^R)-R( \Omega_p).
$
By (\ref{eq:12.8.68}), 
$
R(T')=R(T)+\Phi(\alpha)-\Phi(\alpha_*).
$
Recall that we assumed $ \Phi(\alpha_*) $ is the maximum potential. Hence $ \Phi(\alpha)-\Phi(\alpha_*)<0 $ and consequently $ R(T')<R(T) $. Therefore $ T $ is not a minimum resistance tree among the trees rooted at other states, which is in contrast with the basic assumption about the tree $T$. Hence, the supposition is false and $ \alpha $ is the potential maximizer. We can use the above analysis to show that all action profiles with maximum potential have the same stochastic potential ($ \varphi $) which means all the potential maximizers are stochastically stable.
$~\hfill \square $\\*
In the following we analyze the algorithm's convergence for the general case of non-separable utilities under extra assumptions. 
\begin{assumption} \label{ass5} For any two player $i$ and $j$, if $\Phi(\alpha_2^i,\alpha^{-i})\geq \Phi(\alpha_1^i,\alpha^{-i})$ then $
	u^i(\alpha_2^i,\alpha^{-i})\geq u^j(\alpha_1^i,\alpha^{-i})$.
\end{assumption}
The intuition behind Assumption \ref{ass5} is that the set of agents has some level of homogeneity. Thus, if the potential function is increased by agent $ i $ changing its action from $ \alpha^i_1 $ to $ \alpha^i_2 $ it is as if any other agent $ j $ had at most the same utility increase. 
\begin{assumption} \label{ass555} The potential function $ \Phi $ is non-decreasing.
\end{assumption}
Note that Assumption \ref{ass555}, does not require knowledge of Nash equilibrium. The intuition is that when action profile is changed such that we move towards the Nash equilibrium, the value of potential function will never decrease. 
\begin{theorem}
	Consider a finite potential game in which all players adhere to P-SBLLL with independent revision probability. Let Assumptions \ref{ass1}, \ref{ass2}, \ref{ass3}, \ref{ass5} and \ref{ass555} hold. If an action profile is stochastically stable then it is potential function maximizer.%Suppose for all players, there exists a minimum difference $ R_{min} $ between utility values of two arbitrary actions and $ R_{min}\geq1 $. 
\end{theorem}
\textbf{Proof:} From the theory of resistance trees, the state $ \alpha $ is stochastically stable if and only if there exists a minimum resistance tree $ T $ rooted at $ \alpha $ (see Theorem 3.1 in (\cite{marden})). Recall that a minimum resistance tree rooted at $ \alpha $ has a minimum resistance among the trees rooted at other states. We use contradiction to prove that $ \alpha $ also maximizes the potential function. As the contradiction assumption, suppose the action profile $ \alpha $ does not maximize the potential. Let $ \alpha_* $ be any action profile that maximizes the potential $\Phi$. Since $ T $ is rooted at $ \alpha $, there exists a path from $ \alpha_* $ to $ \alpha $: $ \Omega=\{\alpha_{*} \rightarrow \alpha\} $. We can construct a new tree $ T' $, rooted at $ \alpha_* $ by adding the edges of $ \Omega^R=\{\alpha \rightarrow \alpha_{*} \} $ to $ T $ and removing the edges of $ \Omega $. Hence,
\begin{align}
\label{eq:root}
\begin{aligned}
R(T')=R(T)+R(\Omega^R)-R(\Omega).
\end{aligned}
\end{align}
%Without loss of generality we normalize the potential function to be between 0 and 1, i.e. $ 0 \leq \phi \leq 1 $. 
If the deviator at each edge is only one player, then the algorithm reduces to BLLL. Suppose there exist an edge $ \alpha_q \rightarrow \alpha_{q+1} $ in $ \Omega^R $ with multiple deviators. The set of deviators is denoted by $ S $. If we show $R(\Omega^R)-R(\Omega)\leq 0$ then $T$ is not the minimum resistance tree and therefore, supposition is false and $\alpha$ is actually the potential maximizer. Note that since players' utilities may not be separable, the utility of each player depends on the actions of other players. \\*
From Lemma \ref{lemnr1}, for any transition $\alpha_1 \rightarrow \alpha_2$ in $\Omega^R$,
$
R(\alpha_1 \rightarrow \alpha_2)= \sum_{i \in S_{12}} \max\allowbreak\{u^i(\alpha_1),u^i(\alpha_2)\}-u^i(\alpha_2), 
$
and
$
R(\alpha_2 \rightarrow \alpha_1)= \sum_{i \in S_{21}} \max\{u^i(\alpha_2),u^i(\alpha_1)\}-u^i(\alpha_1),
$
where $ S_{12} $ is the set of deviating players during the transition $ \alpha_1 \rightarrow \alpha_2 $. By Assumption \ref{ass2}, $ S_{12} = S_{21}=S$. Therefore $R(\alpha_1 \rightarrow \alpha_2) - R(\alpha_2 \rightarrow \alpha_1)=$
\begin{align}
\label{eq:12.8.66.p}
\begin{aligned}
{}& \Big[\sum_{i \in S_{12}} \max\{u^i(\alpha_1),u^i(\alpha_2)\}-u^i(\alpha_2)\Big]-\Big[\sum_{i \in S_{21}} \max\{u^i(\alpha_2),u^i(\alpha_1)\}-u^i(\alpha_1)\Big]=\\
&\sum_{i \in S} \Big[ \max\{u^i(\alpha_1),u^i(\alpha_2)\}-u^i(\alpha_2)  - \max\{u^i(\alpha_2),u^i(\alpha_1)\}+u^i(\alpha_1)\Big].
\end{aligned}
\end{align}
By canceling identical terms $max\{u^i(\alpha_2),u^i(\alpha_1)\}$,
\begin{align}
\label{eq:12.8.66.12.p}
\begin{aligned}
{}&R(\alpha_1 \rightarrow \alpha_2) - R(\alpha_2 \rightarrow \alpha_1)= \sum_{i \in S} u^i(\alpha_1)-u^i(\alpha_2).
\end{aligned}
\end{align}
Now divide $\alpha_1 \rightarrow \alpha_2$ into a sequence of sub-edges in which only one agent is deviating. Assuming $ S=\{i,i+1,i+2,...,i+|S|\} $ as the set of deviating players for this transition, the expanded from of this edge can be written as
$$
\alpha_1 \rightarrow \alpha_{2} = \{\alpha_1=\alpha_{k_{0}} \rightarrow (\alpha_2^i,\alpha_1^{-i})=\alpha_{k_1} \rightarrow \alpha_{k_2} \rightarrow ... \rightarrow \alpha_{k_{q-1}} \rightarrow \alpha_{k_q}=\alpha_{2} \},
$$
where $ \alpha_{k_{j}} \rightarrow \alpha_{k_{j+1}} $ is a sub-edge in which only player $i+j$th is deviating. Note that $ q=|S| $. We can rewrite (\ref{eq:12.8.66.12.p}) as
\begin{align}
\label{eq:12.8.66.12.p.5}
\begin{aligned}
{}&R(\alpha_1 \rightarrow \alpha_2) - R(\alpha_2 \rightarrow \alpha_1)= \sum_{i \in S} \mathbold{u^i(\alpha_1)}-u^i(\alpha_{k_1})+u^i(\alpha_{k_1})-
u^{i+1}(\alpha_{k_2})+\\
&u^{i+1}(\alpha_{k_2})
-...+u^{i+|S|-1}(\alpha_{k_{q-1}})-\mathbold{u^i(\alpha_2)}.
\end{aligned}
\end{align}
By Assumption \ref{ass555} and by knowing that $\alpha_1 \rightarrow \alpha_2$ is in $\Omega^R=\{\alpha \rightarrow \alpha_* \}$:
$
\Phi(\alpha_1)=\Phi(\alpha_{k_{0}}) \leq \Phi(\alpha_{k_{1}})
\leq \Phi(\alpha_{k_{2}}) \leq ... \leq \Phi(\alpha_{k_{q}})=\Phi(\alpha_2). 
$
Thus, by Assumption \ref{ass5}:
$
\mathbold{u^i(\alpha_1)}-u^i(\alpha_{k_1}) \leq 0, u^i(\alpha_{k_1})-
u^{i+1}(\alpha_{k_2}) \leq 0, ... , u^{i+|S|-1}(\alpha_{k_{q-1}})-\mathbold{u^i(\alpha_2)} \leq 0
$. 

This means that $R(\alpha_1 \rightarrow \alpha_2) - R(\alpha_2 \rightarrow \alpha_1) \leq 0$. Since $\alpha_1 \rightarrow \alpha_2$ is an arbitrary transition in $\Omega^R$ and $\alpha_2 \rightarrow \alpha_1$ is its reverse transition in $\Omega$, it is obvious that
$
R(\Omega^R)-R(\Omega) \leq 0.
$
Therefore, from (\ref{eq:root}), $T$ is not a minimum resistance tree among the trees rooted at other states, which is in contrast with the basic assumption about $ T $. Hence, the supposition is false and $ \alpha $ is the potential maximizer.
%According to (\ref{eq:12.8.60}) the maximum value for $ R(a_q \rightarrow a_{q+1}) $ is $ |S|\phi_{max} $, i.e. $ R(a_q \rightarrow a_{q+1}) \leq |S|\phi_{max} $. On the other hand the minimum value for $ R(a_{q+1} \rightarrow a_{q}) $ in $ \Omega $ is $ |S| R_{min} $. Therefore by adding a multi-deviation edge we obtain:
%$$
%R(T')=R(T)+|S|\phi_{max}-|S|R_{min}
%$$
%Since $ R_{min} \geq 1 $ we obtain $ |S|-|S|R_{min} \leq 0 $. Therefore $ T $ is not the minimum resistance tree which is in contrast with the assumption about $T$. Hence, the supposition is false and $ a $ is also the potential maximizer.
%Now assume we have found another edge with multiple deviators. With the same approach, the above discussion is repeatable for the new graph. Hence, the stochastically stable action profile $ a $ remains as the potential function maximizer as we add the edges with multiple deviators.
$~ \hfill \square $ 

\subsection{Model-based P-SBLLL}
Note that in P-SBLLL's (\ref{eq:12.8.4n}) and (\ref{eq:12.8.5n}), player $i$ needs to know the utility value of any action $\beta \in \mathcal{A}^i$. However, in the case of unknown environment, utility values of actions in $\mathcal{A}^i$ that are not played are not available to player $i$.

In a model-based learning algorithm, we assume that players can eventually form a belief of the environment by constantly interacting with the environment. This belief is called model. By using the model, when the environment is unknown, players are able to estimate the payoff of the actions that they did not play. In a model-based learning, each player $ i $ possesses an unbiased and bounded estimate $ \hat{u}_\alpha^i $ for the payoff of each action $ \alpha \in \mathcal{A}^i $ (including the actions that he did not play) such that
$
\mathds{E}[\hat{u}_\alpha^i(n+1)|H_n]={u}_\alpha^i(n),
$
where $ H_n $ is the history of the process up to the iteration $ n $. Thus, in a model-based LLL scheme, utility values of unknown actions is replaced by player's estimation model.

The model is usually a function that is fitted over the utility values of the actions that are already sampled. Therefore, one of the main assumptions in model-based learning is that players know the structure of the utility function so that they can fit a function accordingly. 

We conclude this section by emphasizing that P-SBLLL is a type of learning algorithm in which each agent's action set is constrained (recall the completeness assumption). Furthermore, unlike BLLL in which only one random agent takes a trial action, all the agents are allowed to explore the environment. Another valuable feature of P-SBLLL is that each agent's exploration process can be precisely regulated by the revision probability. Thus, each agent can decide independently, and not randomly, whether it is the time for taking a trial action or it is better to remain on the current state. More interestingly, the revision probability can be conditioned on each agent's situation. This will certainly reduce redundant explorations in P-SBLLL comparing to BLLL.

Despite the above-mentioned improvements, P-SBLLL still relies on an estimation model of the environment.~This means that players in P-SBLLL have a prior knowledge about the structure of the utility distribution so they can apply an appropriate model on the environment. In practice, this may not be an appropriate assumption. In order to relax this assumption, we turn to RL algorithms which do not require a model of the environment.

\section{Second Order Q-learning}
\label{soql}
%Recall that in model-based LLL, players need to know the structure of the utility function while RL is model free and players only need to monitor the sequence of their own actions' outcome. Thus, in a MAS setup, where the utility structure is unknown RL can be employed. However, due to the lack of model from the environment, RL needs more time to explore it. 
Q-learning, as a sub-class of RL algorithms, is widely used for policy synthesis problems. In Q-learning algorithm the actions' expected utilities can be compared without requiring a knowledge of the utility function. Thus, in situations where the utility structure is unknown and a model-based LLL can not be used, we may use Q-learning. Furthermore, Q-learning can be used in processes with stochastic transitions, without requiring any adaptations.

Recall that RL dynamics typically uses first-order aggregation where actions' payoff is used to determine the growth rate of the players' score variable
$
{\dot{p}}_\beta^i=u^i(\beta,\alpha^{-i})-T p_\beta^i.
$
We now discuss what happens beyond first-order, when using payoffs
as higher-order forces of change. To that end we first introduce the notion of strictly dominated strategy and weakly dominated strategy:
\begin{definition}
	A strategy $ \alpha \in \mathcal{A}^i $ is strictly (weakly) dominated by strategy $ \beta \in \mathcal{A}^i $ if for all $ x^{-i} \in \mathcal{X}^{-i} $ we have $ u^i(\alpha,x^{-i}) < u(\beta,x^{-i}) $ ($ u^i(\alpha,x^{-i}) \leq u(\beta,x^{-i}) $).
\end{definition}
For a pure strategy $ \beta \in \mathcal{A}^i$, extinction means that $ x_\beta^i \rightarrow 0 $ as $ t \rightarrow \infty $. Alternatively, a mixed strategy $ q^i \in \mathcal{X}^i $ becomes extinct along $ x(t) $ if $ D_{kl}(q^i || x^i)=\sum_\beta q_\beta^i \log(q_\beta^i/x_\beta^i) \rightarrow \infty $ as $ t \rightarrow \infty $. 
\vspace{1mm}

The evolution of the system's dynamic and the efficiency of players' learning crucially depend on how players aggregate their past observation and update their score variable. With this in mind we discuss an $ n $-fold, i.e. high-order, aggregation scheme in RL:
$
{p_\beta^i}^{(n)} (t)=u^i(\beta,\alpha^{-i}) 
$
where ${p_\beta^i}^{(n)}$ denotes $n$th-order derivative. Needless to say, an $ n $th-order aggregation scheme looks deeper into the past observations. Theorem 4.1 in (\cite{num2}) proves that in a high-order continuous-time learning, dominated strategies become extinct. Furthermore, Theorem 4.3 in (\cite{num2}) shows that unlike the first-order dynamics in which weakly dominated strategies may survive, in high-order dynamics ($ n \geq 2 $) weakly dominated strategies become extinct.

In this section we propose a discrete-time Q-learning algorithm in which we incorporated a second-order reinforcement in the aggregation step. Such high-order reinforcement provides more momentum towards the strategies that tend to perform better. We call this algorithm Second-Order Q-Learning (SOQL).

Recall that in a RL algorithm, the score variable characterizes the relative utility of a particular action. In the Q-learning algorithm, the score variable is represented by the Q-value. The Q-value corresponding to the selected action will be updated by the generated reinforcement signal once that action is performed. In standard Q-learning, the update rule for Q-value is 
$$
Q_\beta^i(n+1)=Q_\beta^i(n)+\mu(n)~
[u_\beta^i(n)-Q_\beta^i(n)],
$$
where $ u_\beta^i(n)=\mathds{1}_{\{\alpha^i(n)=\beta\}}
~u^i(n)$ and $ u^i(n) $ is the player $ i $'s realized utility at time step $ n $,  $ Q_\beta^i(n) $ is the player $ i $'s Q-value corresponding to action $ \beta $, and $ \mu(n) $ is the step size assumed to be diminishing. In the action selection step, each player selects his action according to a Q-value based SBR:
\begin{align}
\label{eq:16.999}
\begin{aligned}
X^i_\beta(n+1)=\frac{\exp({1/\tau} Q_\beta^i(n))}{\sum_{\beta' \in \mathcal{A}^i} \exp({1/\tau} Q_{\beta'}^i(n))}.
\end{aligned}
\end{align}

In SOQL we propose a double aggregation process for updating Q-value. Furthermore, to increase SOQL's convergence rate we also propose to use an update rule with constant step size $ \mu $:
\begin{align}
\label{eq:17}
\begin{aligned}
{}\
& P_\beta^i(n+1)=P_\beta^i(n)+\mu~[u_\beta^i(n)-P_\beta^i(n)],\\
& Q_\beta^i(n+1)=Q_\beta^i(n)+\mu~[P_\beta^i(n)-Q_\beta^i(n)],
\end{aligned}
\end{align}
For its mixed strategy, player $i$ uses a greedy action selection similar to (\cite{yatao}):
\begin{align}
\label{eq:18}
\begin{aligned}
X^i(n+1)=(1-\vartheta)X^i(n)+\vartheta BR^i(Q^i(n)),
\end{aligned}
\end{align}
where the constant coefficient $ \vartheta $ is the action selection step size satisfying $ \vartheta<\mu<1 $ and player $i$'s best-response correspondence $BR^i(Q^i(n))$ is defined as
$
BR^i(Q^i(n))=\{e^i_{\alpha_*^i}~|~\alpha_*^i \in \mathcal{A}^i,~ Q_{\alpha_*^i}^i(n) = \max_{\alpha \in \mathcal{A}^i}~Q_\alpha^i(n)\}
$
where $e^i_{\alpha_*^i}$ is player $i$'s pure strategy or the unit vector corresponding to $\alpha_*^i$. 

\subsection{SOQL's Convergence Analysis}
In the following we give conditions under which our proposed Q-learning algorithm converges to a pure strategy Nash equilibrium almost surely.
\begin{proposition} 
	\label{propql1}
	If an action profile $ \alpha(n)=(\beta,\alpha^{-i}) $ is repeatedly played for $ m>0 $ iterations, i.e. $ \alpha(n+c)=\alpha(n) $ for all $ 1 \leq c < m $, then:
	\begin{align}
	\label{eq:19}
	\begin{aligned}
	&P_\beta^i(n+m)=(1-\mu)^m P_\beta^i(n)+(1-(1-\mu)^m)u_\beta^i(n),
	\end{aligned}
	\end{align}
	and
	\begin{align}
	\label{eq:20}
	\begin{aligned}
	&Q_\beta^i(n+m)=m(1-\mu)^{m-1} Q_\beta^i(n+1)-(m-1)(1-\mu)^m Q_\beta^i(n)\\
	&~~~~~~~~~~+\big[(m-1)(1-\mu)^m-m(1-\mu)^{m-1}+1\big]~u_\beta^i(n).
	\end{aligned}
	\end{align}
\end{proposition} 
\textbf{Proof:} From recursively using (\ref{eq:17}) it is easy to show that (\ref{eq:19}) and (\ref{eq:20}) hold.
$~ \hfill \square $ 
\begin{corollary} 
	\label{coroql1}
	For sufficiently large $ m $ if $ 0<\mu<1 $ we have $
	lim_{m \rightarrow \infty} ~Q_\beta^i (n+m)=u_\beta^i(n).
	$
\end{corollary}
\textbf{Proof:} Corollary \ref{coroql1} can be easily verified by taking the limit $ m \rightarrow \infty $ in (\ref{eq:20}), i.e. 
$
lim_{m \rightarrow \infty} m(1-\mu)^{m-1}=0,~
$
$
lim_{m \rightarrow \infty} (m-1)(1-\mu)^{m}=0,~
$
$
lim_{m \rightarrow \infty} (m-1)(1-\mu)^{m}-m(1-\mu)^{m-1}+1=1.
$
$~ \hfill \square $ \\*
\noindent
%\begin{remark} 
%	\label{rem}
%	Note that in Corollary \ref{coroql1}, during the $ m $ iterations, other actions $\beta' \neq \beta,~ \beta' \in \mathcal{A}^i$ are never played. In other words $ Q_{\beta'}^i(n+m)=Q_{\beta'}^i(n) $.
%\end{remark}
\begin{proposition} 
	\label{propql2}
	Let $ \alpha_*=(\beta_*,\alpha_*^{-i}) $ be the action profile that is played at time step $ n $ and $n+1$ where $ e^i_{\beta_*}=BR^i(Q^i(n)) $. We show that if $ 0<\mu<1 $ then at any following $ m $th iteration, with the probability of at least $\prod_{c=1}^{m} \big(1-(1-\vartheta)^c\big)^N$, 
	\begin{align}
	\label{eq:22}
	\begin{aligned}
	& Q_{\beta_*}^i(n+m+1)=(m+1)(1-\mu)^{m} Q_{\beta_*}^i(n+1)-(m)(1-\mu)^{m+1} Q_{\beta_*}^i(n)\\
	&~~~~~~~~~~~~~~~~+\big[m(1-\mu)^{m+1}-(m+1)(1-\mu)^{m}+1\big]~u_{\beta_*}^i(n),
	\end{aligned}
	\end{align}
	and
	\begin{align}
	\label{eq:23}
	\begin{aligned}
	& X^i(n+m+1)=(1-\vartheta)^{m+1}~X^i(n)+(1-(1-\vartheta)^{m+1}) e^i_{\beta_*}.
	\end{aligned}
	\end{align}
\end{proposition} 
\textbf{Proof:} Since $ \beta_* $ is $ BR^i(Q^i(n)) $ and also in Q-learning, Q-values are iteratively converging to $ u_{\beta_*}^i $, for every player $ i \in \mathcal{I} $ and for every action $ \beta \in \mathcal{A}^i,~\beta \neq \beta_* $ we have $  Q_{\beta_*}^i(n+1)>Q_{\beta_*}^i(n)>Q_{\beta}^i(n) $ and $u_{\beta_*}^i(n+1)=u_{\beta_*}^i(n)>Q_{\beta}^i(n)$. We use induction to prove our claim. For $ m=1 $ since $Q_{\beta_*}^i(n+1)>Q_{\beta}^i(n)=Q_{\beta}^i(n)$ it follows that at time step $n+2$, $\alpha_*$ is still the estimated best response. By (\ref{eq:18}) and considering the fact that $BR^i(Q^i(n))=e^i_{\beta_*}$ then at time $n+2$ we have 
$
X^i(n+2)=(1-\vartheta) X^i(n+1)+\vartheta e^i_{\beta_*}.
$
Since $ e^i_{\beta_*} $ is a unit vector, $ X^i_{\beta_*}(t)> \vartheta $ and it is true for all other players. Therefore, at time step $ n+2 $, the action profile $ \alpha_* $ is played with the probability of at least $ \vartheta^N$. From the assumption that for player $ i $, $ u_{\beta_*}^i(n)>Q_{\beta_*}^i(n) $, we have:
\begin{align}
\label{eq:24}
\begin{aligned}
\mu^2 u_{\beta_*}^i(n)> \mu^2 Q_{\beta_*}^i(n).
\end{aligned}
\end{align}
Now, consider the condition $ Q_{\beta_*}^i(n+1)>Q_{\beta_*}^i(n) $. For $ 0<\mu<1 $ it is easy to show that 
\begin{align}
\label{eq:25}
\begin{aligned}
2(1-\mu) Q_{\beta_*}^i(n+1)>2(1-\mu) Q_{\beta_*}^i(n).
\end{aligned}
\end{align}
By combining (\ref{eq:24}) and (\ref{eq:25})
$$
2(1-\mu) Q_{\beta_*}^i(n+1)+\mu^2 u_{\beta_*}^i(n)>2(1-\mu) Q_{\beta_*}^i(n)+\mu^2 Q_{\beta_*}^i(n).
$$
By subtracting $(1-\mu)^2 Q^i_{\beta_*}(n)$ from the both sides,
\begin{align}
\label{eq:26}
\begin{aligned}
2(1-\mu) Q_{\beta_*}^i(n+1)+\mu^2 u_{\beta_*}^i(n)-(1-\mu)^{2} Q_{\beta_*}^i(n)>\\
2(1-\mu) Q_{\beta_*}^i(n)+\mu^2 Q_{\beta_*}^i(n)-(1-\mu)^{2} Q_{\beta_*}^i(n).
\end{aligned}
\end{align}
From (\ref{eq:20}), $2(1-\mu) Q_{\beta_*}^i(n+1)+\mu^2 u_{\beta_*}^i(n)-(1-\mu)^{2} Q_{\beta_*}^i(n)$ is equivalent to $Q_{\beta_*}^i(n+2)$ and therefore:
\begin{align}
\label{eq:26.pp}
\begin{aligned}
Q_{\beta_*}^i(n+2)>Q_{\beta_*}^i(n)>Q_{\beta}^i(n)
\end{aligned}
\end{align}
Recall that other actions $ \beta \in \mathcal{A}^i $ are not played during the $ m $ iterations, i.e. $ Q_{\beta}^i(n)=Q_{\beta}^i(n+1)=Q_{\beta}^i(n+2) $. By substituting $ Q_{\beta}^i(n) $ with $Q_{\beta}^i(n+2)$ in (\ref{eq:26.pp}) we get $ Q_{\beta_*}^i(n+2)>Q_{\beta}^i(n+2) $ which means that $ \beta_* $ is still the estimated best response for player $ i $ at iteration $ n+2 $. By repeating the argument above for all players we show that at time step $ n+2 $, $ \alpha_* $ remains the estimated best response and the claim in (\ref{eq:22}) follows for $ m=1 $.

Now assume, at every iteration $ c $ where $ 1 \leq c \leq m-1 $, $ a_* $ is played with probability $ \prod_{c=1}^{m-1} \big(1-(1-\vartheta)^c\big)^N $. At time step $m$, $X^i$ is updated as: 
\begin{align}
\label{eq:28}
\begin{aligned}
X^i(n+m+1)=(1-\vartheta)^{m+1}~X^i(n)+
(1-(1-\vartheta)^{m+1})e^i_{\beta_*}.
\end{aligned}
\end{align}
By Proposition \ref{propql1}, 
\begin{align}
\label{eq:22.np}
\begin{aligned}
& Q_{\beta_*}^i(n+m+1)=(m+1)(1-\mu)^{m} Q_{\beta_*}^i(n+1)-(m)(1-\mu)^{m+1} Q_{\beta_*}^i(n)\\
&~~~~~~~~~~~~~~~~+\big[m(1-\mu)^{m+1}-(m+1)(1-\mu)^{m}+1\big]~u_{\beta_*}^i(n),
\end{aligned}
\end{align}
Since for player $ i $ any action $ \beta \neq \beta_* $ is not played, it follows that $Q_{\beta}^i(n+m+1)=Q_{\beta}^i(n)$. Moreover, $ Q_{\beta_*}^i(n+m+1)>Q_{\beta_*}^i(n+m)>Q_{\beta}^i(n+m) $ and $ u_{\beta_*}^i(n+m+1)>Q_{\beta}^i(n+m) $.
From (\ref{eq:28}), at time step $n+m+2$, $\alpha_*$ with probability of at least $\big(1-(1-\vartheta)^n\big)^N$ and from Proposition \ref{propql1}, 
\begin{align}
\label{eq:20.np}
\begin{aligned}
Q_\beta^i(n+m+2)\!=\!2(1-\mu) Q_\beta^i(n+m+1)\!-\!(1-\mu)^2 Q_\beta^i(n+m)+\mu^2 u_\beta^i(n+m).
\end{aligned}
\end{align}
With the same argument as for $n=1$,
$
Q_{\beta_*}^i(n+m+2)>Q_{\beta}^i(n+m+2).
$
Therefore, the estimated best response for player $ i $ is not changed. In other words
$
BR^i(Q^i(n+m+2))=BR^i(Q^i(n))=e^i_{\beta_*}.
$
By substituting (\ref{eq:28}) into (\ref{eq:18}) we get $
X^i(n+m+2)=(1-\vartheta)^{m+2}X^i(t)+
(1-(1-\vartheta)^{m+2}) e^i_{\beta_*},
$
which completes the induction argument.

%By Proposition \ref{propql2} it is easy to show 
%\begin{align}
%\label{eq:27}
%\begin{aligned}
%Q_{\beta_*}^i(n+m+2)=2(1-\mu) Q_{\beta_*}^i(n+m+1)-(1-\mu)^{2} Q_{\beta_*}^i(n+m)+\mu ^2 ~u_{\beta_*}^i(n).
%\end{aligned}
%\end{align}
%Since for player $ i $ any action $ \beta \neq \beta_* $ is not played, it follows that $Q_{\beta}^i(n+m)=Q_{\beta}^i(n)$ (Remark \addtocounter{remark}{-1}{\arabic{remark}}\addtocounter{remark}{1}). Moreover, $ Q_{\beta_*}^i(n+m+1)>Q_{\beta_*}^i(n+m)>Q_{\beta}^i(n+m) $ and $ u_{\beta_*}^i(n+m)>Q_{\beta}^i(n+m) $. By using the same argument in (\ref{eq:25}) and (\ref{eq:26}) we can show that 
%$$
%Q_{\beta_*}^i(n+m+2)>Q_{\beta}^i(n+m+2).
%$$
%Therefore, the estimated best response for player $ i $ is not changed. In other words
%$$
%BR^i(Q^i(n+m+2))=BR^i(Q^i(n))=e^i_{\beta_*}.
%$$
%By repeatedly substituting (\ref{eq:18}) into itself,
%\begin{align}
%\label{eq:28}
%\begin{aligned}
%X^i(n+m+1)=(1-\vartheta)^{m+1}~X^i(n)+
%(1-(1-\vartheta)^{m+1}) [\dfrac{exp(u_{\beta_*}^i(n))}{exp((1/N)\Sigma_{i=1}^N u_{\alpha_*^i}^i(n))}] ~e^i_{\beta_*}.
%\end{aligned}
%\end{align}
%By substituting (\ref{eq:28}) into (\ref{eq:18}) and 
%setting $ BR^i(Q^i(n+m+1))=e^i_{\beta_*}$
%\begin{align}
%\label{eq:28.5}
%\begin{aligned}
%X^i(n+m+2)=(1-\vartheta)^{m+2}~X^i(n)+
%(1-(1-\vartheta)^{m+2}) [\dfrac{exp(u_{\beta_*}^i(n))}{exp((1/N)\Sigma_{i=1}^N u_{\alpha_*^i}^i(n))}] ~e^i_{\beta_*}.
%\end{aligned}
%\end{align}
%Following the same argument as for $ m=1 $, it follows that $ \alpha_* $ is played with probability of at least $ (1-(1-\vartheta)^{m+2})^N$. This completes the induction argument.
$~ \hfill \square $ 
\begin{corollary} 
	\label{coroql2}
	If for sufficiently large $ m>0 $ the conditions of Proposition \ref{propql2} hold, then for every player $ i $ the following holds with probability $\prod_{c=1}^{\infty} [\big(1-(1-\vartheta)^c\big)^N$:
	\begin{align}
	\label{eq:29}
	\begin{aligned}
	\lim_{m \rightarrow \infty} ~Q_{\beta_*}^i(n+m)=u^i(\alpha_*),~
	\lim_{m \rightarrow \infty} ~X^i(n+m)=e^i_{\beta_*}.
	\end{aligned}
	\end{align}
\end{corollary}
\begin{theorem} 
	\label{soqlth}
	For sufficiently large $ m $, if the conditions in Proposition \ref{propql2} hold, then $ X(n+m) $ converges to a neighborhood of $ \alpha_* $ with probability one.
\end{theorem}
\textbf{Proof:} From Proposition \ref{propql2} we know that $ \alpha_* $ is played with probability $\prod_{c=1}^{\infty}\allowbreak [\big(1-(1-\vartheta)^c\big)^N$. Inspired by Proposition 6.1 in \cite{chasparis}, we show that this probability is strictly positive. The product $ \prod_{c=1}^{\infty} (1-(1-\vartheta)^c)$ is non-zero if and only if $ \sum_{c=1}^{\infty} \log(1-(1-\vartheta)^c)> - \infty$. Alternatively we can show
\begin{align}
\label{eq:31}
\begin{aligned}
-\sum_{c=1}^{\infty} \log(1-(1-\vartheta)^c)< \infty.
\end{aligned}
\end{align}
By using the limit comparison test and knowing $ 0<1-\vartheta<1 $:
$$
\lim_{c \rightarrow \infty} \dfrac{-\log(1-(1-\vartheta)^c)}{(1-\vartheta)^c}=\lim_{c \rightarrow \infty} \dfrac{1}{1-(1-\vartheta)^c)}=1.
$$
Therefore, (\ref{eq:31}) holds if and only if $ \sum_{c=1}^\infty~(1-\vartheta)^c < \infty $. The latter holds since
$$
\sum_{c=1}^\infty~(1-\vartheta)^c=\dfrac{1}{1-(1-\vartheta)}=\dfrac{1}{\vartheta}~<~\infty.
$$
By (\ref{eq:31}) we have
$
\lim_{m \rightarrow \infty}  \prod_{c=1}^{m} (1-(1-\vartheta)^m)>0,
$
so we can conclude that
$
\forall \eta ,~ 0<\eta<1,~ \exists M ~s.t.~ (1-(1-\vartheta)^m)\geq\eta~~\forall m \geq M.
$
In other words after $ M $ iterations $ X(n+M) $ enters a ball $ B_{1-\eta}(\alpha_*) $ with probability $ \prod_{c=1}^{M} (1-(1-\vartheta)^c)^N $. As discussed in \cite{yatao}, and following the proof of Theorem 3.1 in \cite{marden2}, for $ M \geq \log_{1-\vartheta} (1-\eta)>0 $, trajectories of $ X(n+M) $ enters a neighborhood of $ \alpha_* $ with probability $ \prod_{c=1}^{\infty} (1-(1-\vartheta)^c)^N $; hence, converges to $ \alpha_* $ almost surely.\\
$~ \hfill \square $
\subsubsection{Perturbation}
\label{soqlpert}
We proved that the SOQL converges to a neighborhood of $\alpha_*$, i.e. $B_{1-\eta}(\alpha_*)$, almost surely. Recall that in a learning algorithm it is essential that the action space is well explored; otherwise, it is likely that the estimated equilibrium does not converge to the true Nash equilibrium. Thus, the action selection procedure should allow players to explore new actions even if the new actions are sub-optimal. In this section we discuss the conditions under which the estimated equilibrium can reach an actual Nash equilibrium. The perturbation functions are usually carefully designed to suit this need. A perfect perturbation function is decoupled from
dynamics of the learning algorithm, adjustable and has minimum effect on perturbing
the optimal solution (\cite{yatao}).

In the standard Q-learning scheme, the Boltzmann action selection map has already incorporated the exploration feature by using the temperature parameter $\tau$ (\cite{leslie-collins-2005}). In order to implement such exploration feature in the modified algorithm, we use a perturbation function. The perturbed mixed strategy for player $i$ is defined as
\begin{align}
\label{eq:32}
\begin{aligned}
\tilde{X}_j^i=(1-\rho^i(X^i,\xi))X_j^i+\rho^i(X^i,\xi)~\mathds{1}_j^i / |\mathcal{A}_i|.
\end{aligned}
\end{align}
where $ \rho^i(X^i,\xi) $ is the perturbation function. The perturbation function $ \rho^i: \mathcal{X}^i \times [\bar{\epsilon},1] \rightarrow [0,1]$ is a continuously differentiable function where
for some $ \xi \in (0,1) $ sufficiently close to one, $ \rho^i $ satisfies the following properties:
\begin{itemize}
	\item $ \rho^i(X^i,\xi)=0,~ \forall X^i$ such that $ \forall \xi \geq \bar{\epsilon}:~ |X^i|_\infty<\zeta$,
	\item $ \lim_{|X^i|_\infty \rightarrow 1} ~\rho^i(X^i,\xi)=\xi$,
	\item $ \lim_{|X^i|_\infty \rightarrow 1} \frac{\partial \rho^i(X^i,\xi)}{\partial X_j^i}~|_{\xi=0}=0,~\forall j \in \mathcal{A}^i$.
\end{itemize} 
As in (\ref{eq:32}), each player $i$ selects a random action with a small probability $ \rho_i $ and selects the action with highest $Q$ value with the probability $(1-\rho_i) $.
\begin{assumption} \label{ass6} Step sizes in Q update rule are adjusted based on
	$
	\mu_j^i(n)=(1-\tilde{X}_j^i(n)).
	$
\end{assumption}
\begin{assumption} \label{ass7} When all the players enter the perturbation zone, i.e. $ |X^i|_\infty>\zeta,~\forall i \in \mathcal{I} $, no more than one player chooses a sub-optimal action at each iteration (asynchronous perturbation).\end{assumption} 
\begin{theorem} 
	If the conditions in Proposition \ref{propql2} hold for a sufficiently large $ m $, then under Assumption \ref{ass6} and Assumption \ref{ass7}, the estimated equilibrium $ \alpha_* $ converges to a Nash equilibrium almost surely.
\end{theorem}
\textbf{Proof:} From Corollary \ref{coroql2}
we have $
\lim_{m \rightarrow \infty} ~Q_{\beta_*}^i(n+m)=u^i(\alpha_*),
$
and 
$$
\lim_{m \rightarrow \infty} ~X^i(n+m)= e^i_{\beta_*},
$$
where $\alpha_*=(\beta_*,\alpha_*^{-i})$. Assume that perturbation becomes active at some large time step $ \bar{n}+1 $ and player $ i $ chooses an action $ \beta \neq \beta_* $. From (\ref{eq:32}) we know that such perturbation happens with the probability of at least $ \xi/|\mathcal{A}^i| $. From (\ref{eq:17}) and Assumption \ref{ass6}, player $ i $ updates its action by
\begin{align}
\label{eq:33}
\begin{aligned}
Q_{\beta}^i(\bar{n}+2)=2 \bar{X}_{\beta}^i(\bar{n})Q_{\beta}^i(\bar{n}+1)
-\bar{X}_{\beta}^i(\bar{n})^2Q_{\beta}^i(\bar{n})+ (1-\bar{X}_{\beta}^i(\bar{n}))^2u^i(\beta(\bar{n}),\alpha_*^{-i}(\bar{n})).
\end{aligned}
\end{align}
Consider the following two cases:
\begin{itemize}
	\item $ u^i(\beta,\alpha_*^{-i}(\bar{n}))<u^i(\alpha_*(\bar{n})) $: In this case player $ i $ failed to improve the utility and will stay at the estimated equilibrium $\alpha_*$ almost surely. 
	\item $ u^i(\beta,\alpha_*^{-i}(\bar{n})) = u^i(\alpha_*(\bar{n})) $: With $ \bar{X}_{\beta}^i $ sufficiently close to zero the Q-values for the actions $ \beta $ is updated to a value sufficiently close to $u^i(\beta,\alpha_*^{-i}(\bar{n}))=u^i(\alpha_*(\bar{n}))$. Hence, in the worst case, when $\alpha_*(\bar{n})$ is only played once, the Q-values for the actions $ \beta $ and $\beta_*$ may become equal. Therefore, in the action selection stage, the set of best response may contain two pure actions. Consequently, the mixed strategies for both actions $ \beta $ and $\beta_*$ is updated by (\ref{eq:18}) and player $ i $ chooses one of them randomly in the next time step.
	\item $ u^i(\beta,\alpha_*^{-i}(\bar{n})) > u^i(\alpha_*(\bar{n})) $: In this case player $ i $ found a response that is better than $\alpha_*$. Since the action $\beta $ has small probability $ \bar{X}_{\beta}^i $, i.e., sufficiently close to 0, then $(1-\bar{X}_{\beta}^i)$ is sufficiently close to 1. Thus from (\ref{eq:33}), $ Q_{\beta}^i $ is updated to a value sufficiently close to $ u^i(\beta,\alpha_*^{-i}(\bar{n}))$ and the action profile $ (\beta,\alpha_*^{-i}(\bar{n})) $ becomes the new estimated best response, i.e., $ \alpha_{*}(\bar{n}+1):= (\beta,\alpha_*^{-i}(\bar{n}))$. Note that player's utility is improved by $u^i(\alpha_*(\bar{n}+1))-u^i(\alpha_*(\bar{n}))$, and the potential would also increase by the same amount. Recall that from the finite improvement property, such improvement in potential function is not limitless and will terminate at the potential maximum, i.e. Nash equilibrium in potential games. Hence, the estimated equilibrium $\alpha_*$ will eventually converge to an actual Nash equilibrium.$~ \hfill \square $ 
\end{itemize}
In the following, through a number of numerical experiments, we compare the performances of P-SBLLL and SOQL with those of BLLL and standard QL.

\section{Case Study}
Multi-robot Coverage Control (MCC) algorithms are concerned with the design of rules for coordinating a set of robots' action. The main objective of agents in MCC is to efficiently cover an area. We assume that there is a set of hidden targets in the area and the robots are assumed to find them. Targets are distributed randomly in the area and each target emits a signal in a form of a symmetric Gaussian function which can be sensed by nearby robots. The area is a limited, two-dimensional, rectangular plane over which a probabilistic function is defined that represents the probability of targets' existence (e.g. (\cite{rahili}), (\cite{cp1.5}) and (\cite{cp1.6})). Therefore, the probability of targets' existence at each point is proportional to the cumulative strength of the signals detected at that point. However, there exist different approaches in the literature to model the search area. One approach is to model the environment as a two-dimensional polyhedron over which an event density function is defined to determine targets' distribution (e.g. (\cite{cp1.5})). Alternatively, the area can be divided into Voronoi regions where the objective of agents in this approach is to converge to the centroids of their assigned regions (e.g. (\cite{c3}), (\cite{cn5}) and (\cite{cn6})). 

During the search for the targets, the mobile sensors regularly collect data from the environment by sensing the signal strength at different locations. The robots use the collected data to ``reinforce" their knowledge about the area or to ``model" the environment. This knowledge will further support agents' decision-making process. Clearly, the overall systems' performance is directly related to agents' location in the environment. Thus, the coverage control problem can be reduced to the problem of optimally locating the robots based on data collected from the area. In the following we present the formulation of our MCC setup.
\subsection{Problem Formulation}
\label{sec:formulation}
Consider a finite number of robots (i.e. agents) as a network of mobile sensors scattered in a geographic area. The area is a two-dimensional plane, divided into a $ L \times L $ square grid. Each cell on this grid is a $ 1 \times 1 $ square and the coordinates of its center are $ l=(l_x,l_y) \in \mathcal{L}$, where $ \mathcal{L} $ is defined as the collection of the centroids of all lattices. The robots can only move between these centroids. 

The location of agent $ i \in  \mathcal{I} $ at iteration $n$ is $ \alpha^{i} (n) := (l_x^{i} (n),l_y^{i} (n))$ \nomenclature{$\alpha^{i} (n)$}{location of agent $ i $ at time step $ n $} $\in \mathcal{L} $ and this defines his action at time step $ n $.  Let the set of robot $ i $'s neighbor cells be defined as $ N_R^{i} (\alpha(n)):= \{ l \in \mathcal{L} |~ | \alpha^{i}(n) - l | \leq R\} $ \nomenclature{$N_R^{i} $}{collection of all neighbors of agent $ i $} where $ R $ \nomenclature{$ R $}{neighborhood radius} is the neighborhood radius. The motion of agents is limited to their adjacent lattices and we use $ \mathcal{A}^{i}_c (n)$ to denote this constrained action set for agent $i$ at time step $n$. 

\begin{figure}[!t]
	\centering
	\subfloat[Two dimensional environment and the set of targets]{{\includegraphics[width=0.4\textwidth]{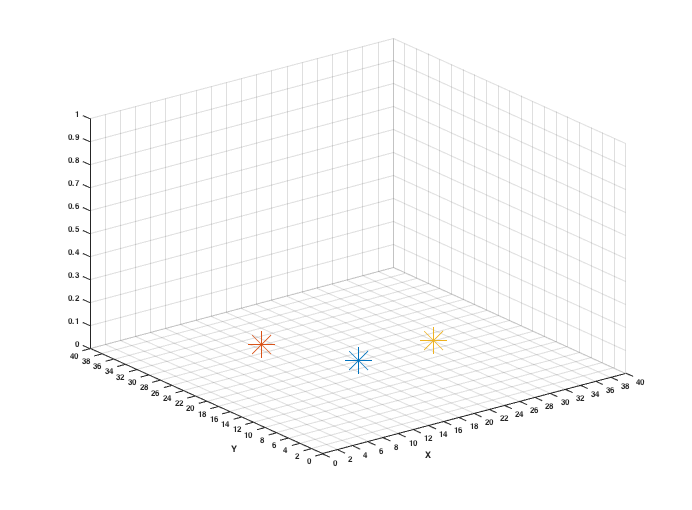}}}
	$~~~~~~~$
	\subfloat[Worth distribution as a Gaussian mixture model]{{\includegraphics[width=0.4\textwidth]{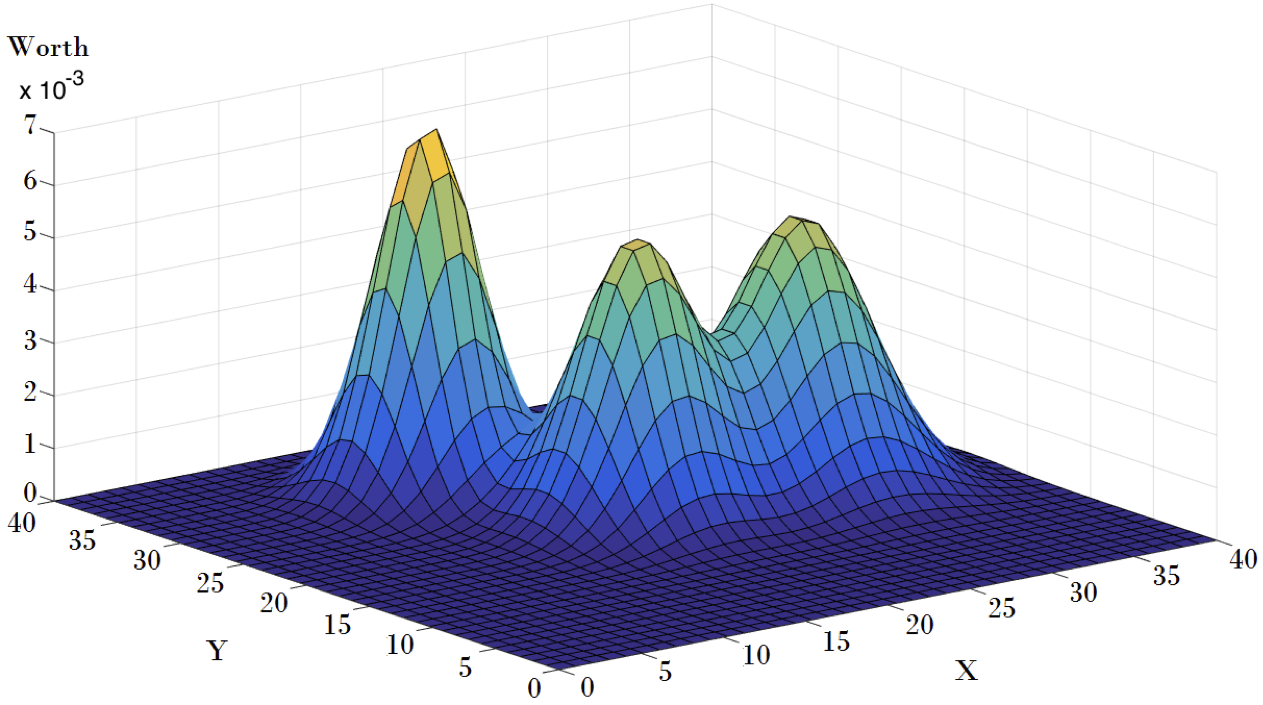} }}%
	\caption{The set of targets and their associated worth distribution}
	\label{f3}
\end{figure}

A number of targets that are distributed in the area (Fig. \ref{f3}.a). Each target emits a signal in the form of symmetric Gaussian function. Therefore, the signal distribution on the area is basically a Gaussian mixture model. The location of the targets is initially unknown for the robots. Thus, the worth of each cell is defined as the probability that a target exists at the center of that cell and is denoted by $ f:~\mathcal{L} \rightarrow [0,1] $\nomenclature{$f(l)$}{the worth at coordinate $ l $}. By sensing targets' signal, the agents can determine $ f(l) $ when they are at the coordinate $ l $. Since there is a direct correlation between the strength of the sensed signal and the worth value at a given cell, the worth distribution is also a Gaussian mixture model (Fig. \ref{f3}.b). The area covered by agent $ i $ at time step $ n $ is $C^i: \mathcal{A}^{i} \mapsto \mathds{R} $ and is defined as $ C^i(\alpha^{i}(n)):=\sum_{{l} \in N_{\delta}^i(n)} f({l}) $ where $\delta$ is the covering range. 

In our setup, each agent $ i $ lays a flag at each cell as he observes that cell; the trace of his flags is denoted by $ \Upsilon^i $ and is detectable by other agents from the maximum distance of $ 2\delta $. This technique allows the agents to have access to the observed areas locally without any need to have access to each other's observation vectors.

\subsubsection{Game Formulation}

An appropriate utility function has to be defined to specify the game's objective and to properly model the interactions between the agents. In our case, we have to define a utility function that takes the reward of sensing worthwhile areas. Inspired by (\cite{rahili}), we also consider the cost of movement of the robots: the movement energy consumption of agent $ i $ from time step $ n-1 $ to $n$ is denoted by $ E^{i}_{move} = K^{i} (| \alpha^{i}(n) - \alpha^{i}(n-1) |)$, where $ K^{i} > 0 $ \nomenclature{$ E^{i}_{move} $}{movement energy consumption of agent $ i $} is a regulating coefficient. The\nomenclature{$ C_{(a(n))} $}{the worth of the covered area by all agents} coverage problem can now be formulated as a game by defining the utility function:
\begin{equation}
\label{eq:1}
\begin{split}
u^{i}(\alpha^i(n),\alpha^{i}(n-1))&=\varrho^{i}[C^i(\alpha^{i}(n))-C^i_n(\alpha^{i}(n))]-K^{i} (| \alpha^{i}(n) - \alpha^{i}(n-1) |), 
\end{split}
\end{equation}
where $ C^i_n(\alpha^{i}(n))=\sum_{j \in \mathcal{I} \setminus i} \sum_{l \in N_{j}^\delta \cap N_{i}^\delta} f({l})$ and $ \varrho^{i} $ is defined as:
\[\varrho^{i} = \left\{
\begin{array}{lr}
1 & : \alpha^{i} \notin \Upsilon^j, j \neq i\\
0 & : otherwise
\end{array}
\right.
\]
Parameter $ \varrho^i $ prevents players to pick their next actions from the areas that are previously observed by other agents. As in (\ref{eq:1}), $C^i(\alpha^{i}(n))-C^i_n(\alpha^{i}(n)) $ is the worth of area that is only covered by agent $ i $. Hence, $u^i$ only depends on the actions of player $i$, i.e. $u^i$ is ``separable" from the actions of other players. 
\begin{lemma} 
	\label{lem2.1} 
	The coverage game $\mathcal{G} := \langle \mathcal{I},\mathcal{A},U_{cov} \rangle$ is a potential game where $ U_{cov} $ is the collection of all utilities $ U_{cov} = \{ u_{i}, i=1,...,N\}$ and the potential function is
	$
	\Phi (\alpha(n),\alpha(n-1))=\sum_{j=1}^N u^{j}(\alpha^j(n),\alpha^{j}(n-1)).
	$
\end{lemma}
\textbf{Proof:} We have to show that there exists a potential function $ \Phi:\mathcal{A} \mapsto \mathds{R} $ such that for any agent $ i \in \mathcal{I}$, every $\alpha^{-i}(n) \in \mathcal{A}^{-i}$ and any $ \alpha^{i}_1(n),\alpha^{i}_2(n) \in \mathcal{A}^{i} $:
\begin{equation}
\label{eq:3.1}
\begin{split}
{}&\Phi (\alpha^{i}_2(n),\alpha^{-i}(n),\alpha(n-1))-\Phi (\alpha^{i}_1(n),\alpha^{-i}(n),\alpha(n-1))=\\
&u^{i}(\alpha^{i}_2(n),\alpha^{-i}(n),\alpha(n-1))-u^{i}(\alpha^{i}_1(n),\alpha^{-i}(n),\alpha(n-1)),
\end{split}
\end{equation}
Using $\Phi$ as in Lemma \ref{lem2.1} we have:
\begin{align*}
\begin{aligned}
{}& \Phi (\alpha^{i}_2(n),\alpha^{-i}(n),\alpha(n-1))= [\Sigma_{j=1,j \neq i}^N u^j(\alpha^j(n),\alpha^j(n-1))]+u^i(\alpha^i_2(n),\alpha^i(n-1))=\\
&\Big[ \Sigma_{j=1,j \neq i}^N \varrho^{j}[C^j(\alpha^{j}(n))-C^j_n(\alpha^{j}(n))]-\Sigma_{j=1,j \neq i}^N[K^{i} (| \alpha^{j}(n)- \alpha^{j}(n-1) |)]\Big] + \\
& \Big[ \varrho^{i}[C^i(\alpha^{i}_2(n))-C^i_n(\alpha^{i}_2(n))]-K^{i} (| \alpha^{i}_2(n)- \alpha^{i}(n-1) |) \Big], \\
\end{aligned}
\end{align*}
and
\begin{align*}
\begin{aligned}
&\Phi (\alpha^{i}_1(n),\alpha^{-i}(n),\alpha(n-1))=[\Sigma_{j=1,j \neq i}^N u^j(\alpha^j(n),\alpha^j(n-1))]+u^i(\alpha^i_1(n),\alpha^i(n-1))=\\
&\Big[ \Sigma_{j=1,j \neq i}^N \varrho^{j}[C^j(\alpha^{j}(n))-C^j_n(\alpha^{j}(n))]-\Sigma_{j=1,j \neq i}^N[K^{i} (| \alpha^{j}(n)- \alpha^{j}(n-1) |)]\Big] + \\
& \Big[ \varrho^{i}[C^i(\alpha^{i}_1(n))-C^i_n(\alpha^{i}_1(n))]-K^{i} (| \alpha^{i}_1(n)- \alpha^{i}(n-1) |) \Big], \\
\end{aligned}
\end{align*}
Hence,
\begin{align*}
\begin{aligned}
&\Phi (\alpha^{i}_2(n),\alpha^{-i}(n),\alpha(n-1))-\Phi (\alpha^{i}_1(n),\alpha^{-i}(n),\alpha(n-1))=\\
&\varrho^{i}[C^i(\alpha^{i}_2(n))-C^i_n(\alpha^{i}_2(n))]-K^{i} (| \alpha^{i}_2(n)- \alpha^{i}(n-1) |)- \varrho^{i}[C^i(\alpha^{i}_1(n))-C^i_n(\alpha^{i}_1(n))]-\\
&K^{i} (| \alpha^{i}_1(n)- \alpha^{i}(n-1) |)= u^{i}(\alpha^{i}_2(n),\alpha^{-i}(n),\alpha(n-1))-u^{i}(\alpha^{i}_1(n),\alpha^{-i}(n),\alpha(n-1)),
\end{aligned}
\end{align*}
and (\ref{eq:3.1}) follows.
$ ~\hfill\square $

\subsubsection{Gaussian Model}
Targets are assumed to emit a signal in a Gaussian form that decays proportionally to the distance from the target; the mobile sensor should be close enough to sense the signal. Thus, the function $f(l)$ is basically a worth distribution in the form of Gaussian Mixture over the environment. A Gaussian Mixture Model ($GMM$) is a weighted sum of Gaussian components (i.e. single Gaussian functions), i.e. $
GMM:=f(l)=\sum_{j=1}^M \omega_{j} ~g(l|\mu_{j},\Sigma_{j}),
$ where $ M $ is the number of components (i.e., targets), $ \omega_{j} $ is the weight (signal strength) of $ j $th component and $
g(l|\mu_{j},\Sigma_{j})=\dfrac{1}{2\pi |\Sigma_{j}| ^{1/2}}~exp[\dfrac{-1}{2}(l-\mu_{j})^T\Sigma_{j}^{-1}(l-\mu_{j})]$ is a two-variable Gaussian function where $ l $ is the location vector; $ \mu_{j} $ is the mean vector (i.e. location of the target $ j $); and $ \Sigma_{j} $ is the covariance matrix of component $ j $. Note that each component represents one target's presence probability and the whole $GMM$ is a mixture of these components. To ensure that the Gaussian distribution is representing the probability map over the area, the summation of all weight coefficients must be equal to one (i.e. $ \sum_{j=1}^M \omega_{j} =1$). $GMM$ parameters can be summarized into the set $ \lambda := \{\omega_{j},\mu_{j},\sigma_{j}|j=1,...,M\} $. 

\subsection{Learning Process}

Recall that a learning scheme can efficiently reach a Nash equilibrium in a potential game based on the game's finite improvement property. This section discusses P-SBLLL and SOQL as learning schemes to solve the MCC problem. We first review P-SBLLL and then we present SOQL's results.
%\begin{figure}[!t] \centering \includegraphics[width=0.7\textwidth]{2QL_FC.png} \caption{The final configuration of agents in SOQL}
%\label{2QL_FC}
%\end{figure}
\subsubsection{P-SBLLL}
A $ 40 \times 40 $ square area with $ 1 \times 1 $ lattices has been considered as the environment. The Gaussian distribution in this area has been chosen randomly with different but reasonable weight, mean and covariance values (Fig.\ref{f3}). The number of targets (Gaussian components) is between 1 to 5 and robots have no prior knowledge about this number. 

A group of five robots ($ N=5 $) are initialized randomly through this $GMM$ to maximize their utility function. We assumed that a robot in our MCC setup is not able to move rapidly from its current lattice to any arbitrary lattice. In other words, the range of its movement is bounded, i.e., constrained. Recall that the P-SBLLL scheme is a modified version of standard LLL which allows players to have a constrained action set in a potential game. Furthermore, we showed that in a potential game, the game will stochastically converge to the potential maximizer if players adhere to P-SBLLL. 

However, in P-SBLLL each agent must have a posteriori knowledge about the utility distribution, i.e. $GMM$, to be able to calculate their future utility function $u^{i}(\alpha^{i}_T,\alpha^{-i}(n),\alpha(n))$ in (\ref{eq:12.8.4n}) and (\ref{eq:12.8.5n}). Hence, in the case of unknown environment, it is not possible to use P-SBLLL scheme, unless we replace $u^{i}(\alpha^{i}_T,\alpha^{-i}(n),\alpha(n))$ with an estimation from the model $\hat{u}^{i}(\alpha^{i}_T,\alpha^{-i}(n),\alpha(n))$. This raises the need for an environmental model in the P-SBLLL algorithm.\\*
\\* 
\noindent
\emph{Unknown Utility Model}:\\*
It is common in practice to assume that the game being played is unknown to the agents, i.e. agents do not have any prior information about the utility function and do not know the expected reward that will result from playing a certain action (\cite{nowe}). It is because the environment may be difficult to assess, and the utility distribution may not be completely known. In the following, inspired by (\cite{rahili}) we introduce an estimation which can provide a model of the environment in model-based learning.\\*
\\*
\noindent
\emph{Expectation Maximization:}
\label{EM_N}
\\*
Assume the environment is a $GMM$. Agent $ i $'s estimation of the mixture parameters $ \lambda := \{\omega_{j},\mu_{j},\sigma_{j}|j=1,...,M\} $ is denoted by $ \hat{\lambda}^i= \{{\hat{\omega}}_{j}^i,{\hat{\mu}}_{j}^i,{\hat{\sigma}}_{j}^i|j=1,...,M\}$. The estimation model is $
\widehat{GMM}:=\hat{f}({l})=\sum_{j=1}^M {\hat{\omega}}_{j} ~g(l|{\hat{\mu}}_{j},{\hat{\Sigma}}_{j}).$ 
During the searching task, the robots will keep their observations of sensed regions in their memories. Let agent $ i $'s observation vector at iteration $ n $ be a sequence of its sensed values from time step $ 1 $ to $ n $ and is denoted by $ O^i=\{O_{1}^i, O_{2}^i, O_{3}^i,...,O_{n}^i\} $ where $ O_{n}^i $ is defined as the corresponding coordinates of the sensed lattice by agent $ i $ at time step $ n $. Expectation Maximization (EM) algorithm is an algorithm that is used in the literature to find maximum likelihood parameters of a statistical model when there are missing data points from the model. The EM algorithm can estimate the $GMM$ parameters $\hat{\lambda}$ through an iterative algorithm (\cite{c8}):
\begin{align}
\label{eq:11}
\begin{aligned}
{}& {\hat{\omega}}_{j}^i= \dfrac{1}{n} \sum_{\tau=1}^n P^i (j | O_{\tau}^i,\hat{\lambda}^i),\\
&{\hat{\mu}}_{j}^i=\dfrac{\sum_{\tau=1}^n P^i (j | O_{\tau}^i,\hat{\lambda}^i)O_{\tau}^i}{\sum_{\tau=1}^n P^i (j | O_{\tau}^i,\hat{\lambda}^i)},\\
&{{\hat{\sigma}}_{j}^{i}}~{}^2=\dfrac{\sum_{\tau=1}^n P^i (j | O_{\tau}^i,\hat{\lambda}^i)(O_{\tau}^i-\hat{\mu_{j}^i})(O_{\tau}^i-\hat{\mu_{j}^i})^T}{\sum_{\tau=1}^n P^i (j | O_{\tau}^i,\hat{\lambda}^i)},\\
&P^i (j | O_{\tau}^i,\hat{\lambda}^i)=\dfrac{\hat{\omega_{j}^i} ~g(O_{\tau}^i|\hat{\mu_{j}^i},\hat{\sigma_{j}^i})}{\sum_{k=1}^M \hat{\omega_{k}^i} ~g(O_{\tau}^i|\hat{\mu_{k}^i},\hat{\sigma_{k}^i})},
\end{aligned}
\end{align}
where $ P^i (j | O_{\tau}^i,\hat{\lambda}^i) $ is often referred to as posteriori probability of the observation vector $O_{\tau}^i$ of agent $ i $ for the $ j $th component of Gaussian distribution. Through (\ref{eq:11}), given the current estimated parameters, the EM algorithm estimates the likelihood that each data point belongs to each component. Next, the algorithm maximizes the likelihood to find new parameters of the distribution. 

Recall that $O^i$ is the sequence of agent $i$'s observed coordinates within the area. Thus, (\ref{eq:11}) only considers the coordinates of the observed lattice $ l $ and disregards its corresponding signal strength $ f(l) $. This issue is pointed out in (\cite{rahili}) and the proposed solution as introduced in (\cite{c9}) is to repeat the iterative algorithm for $ m $ times in worthwhile areas and $ m $ is chosen as:
\[m = \left\{
\begin{array}{lr}
1+V ~round(\dfrac{f(l)}{f_{mode}}) & : f(l)\geq f_{mode}\\
1 & : f(l) < f_{mode}
\end{array}
\right.
\]
where $ f(l) $ is the worth at coordinate of $ l $, $ f_{mode} $ is the threshold above which the signal is considered worthwhile for EM repetition and $ V $ is the correction factor which regulates the number of algorithm repetitions for the worthy lattices. The algorithm will run once for areas with a worth value of less than the threshold.

The variation rate of the estimation parameters $ \hat{\lambda}^i $ is relatively low due to the rate of update of the observation vector and the nature of the EM algorithm. The case of slow variation of Gaussian distribution is investigated in (\cite{c10}). With assuming a slow changing distribution, (\cite{c10}) showed that the agents' estimation error $ |\hat{\lambda}(n)-\lambda(n)| $ is decreased by extending the observations vector; consequently, the agents will stochastically converge to a Nash equilibrium. If we assume that the number of the targets ($ M $) is a known parameter, we can use the iterative algorithm discussed in this section. However, we assumed the agents do not have any prior knowledge about the environment. This lack of knowledge includes the probability distribution function and also the number of targets. Thus, the standard EM algorithm can not be used.

\subsubsection{EM with Split and Merge}
\label{EM_SM}
In this section, we present a new modified EM algorithm that does not need the number of components to calculate $GMM$'s parameters. To achieve this goal, we propose to have a mechanism that can estimate the number of the targets in parallel to parameters estimation. 

The Akaike information criterion (AIC) introduced in (\cite{c11}) is considered as the main verification method to help agents select the best estimation for the number of the targets. The Akaike information criterion is a measure of the relative quality of a distribution model for a set of data points. Given a set of models for the same data, AIC estimates the quality of each model, relative to other statistical models. The AIC value of each model is $
AIC=2k-2ln(L)$ where $ L $ is the maximized value of the likelihood and k is the number of estimated parameters in the model\nomenclature{$AIC$}{Akaike Information Criterion}. Given a collection of statistical models, the one with minimum AIC value is the best model (\cite{c11}).  Akaike criterion considers the goodness of the fit as the likelihood function $ L $ and penalty of complexity of the model as $ k $ (i.e. the number of model parameters). 

Let the estimated number of the targets at time step $n$ each robot $i$ be denoted by $\hat{M}^i(n)$. In our proposed algorithm, after each $ n_{AIC} $ time steps, each agent $i \in \mathcal{I}$ randomly picks a number $T_{AIC}$ from the set $\mathcal{M}(n_{AIC})=\{\hat{M}^i(n_{AIC})+1,\hat{M}^i(n_{AIC})-1\} \subset \mathds{N}$. If $\hat{M}^i(n_{AIC})-1=0$ then $\mathcal{M}(n_{AIC})$ reduces to $\{\hat{M}^i(n_{AIC})+1\}$. Next, player $i$ decides between its current estimated Gaussian component number $\hat{M}^i(n_{AIC})$ and $T_{AIC}$ according to the following probabilities:
\begingroup\makeatletter\def\f@size{8.5}\check@mathfonts
\begin{align}
\label{eq:7.89}
\begin{aligned}
{}&P^{i}_{\hat{M}^i(n_{AIC})}= \dfrac{\exp(\dfrac{1}{\tau} IAIC(\hat{M}^i(n_{AIC})))}{\exp(\dfrac{1}{\tau} IAIC(\hat{M}^i(n_{AIC})))+\exp(\dfrac{1}{\tau} IAIC(T_{AIC}))},
\end{aligned}
\end{align}
\endgroup
\begingroup\makeatletter\def\f@size{8.5}\check@mathfonts
\begin{align}
\label{eq:7.90}
\begin{aligned}
{}&P^{i}_{T_{AIC}}= \dfrac{\exp(\dfrac{1}{\tau} IAIC(T_{AIC}))}{\exp(\dfrac{1}{\tau} IAIC(\hat{M}^i(n_{AIC})))+\exp(\dfrac{1}{\tau} IAIC(T_{AIC}))},
\end{aligned}
\end{align}
\endgroup
where $P^{i}_{\hat{M}^i(n_{AIC})}$ is the probability that player $i$ keeps its current estimation $\hat{M}^i(n_{AIC})$ and $P^{i}_{T_{AIC}}$ is the probability that player $i$ chooses $T_{AIC}$ as its estimation of the number of the components. $IAIC(\hat{M}^i(n_{AIC}))$ is the inverse AIC value for agent $i$'s estimation distribution when agent $i$'s estimated number of the components is $\hat{M}^i(n_{AIC})$ and $IAIC(T_{AIC})$ is the inverse AIC value when agent $i$'s estimation is $T_{AIC}$.

Since the number of the components are changing every $ n_{AIC} $ iterations, the next step is to determine an appropriate method for merging and splitting Gaussian components. The method proposed in (\cite{c1}) incorporates the split and merge operations into the EM algorithm for Gaussian mixture estimations. Moreover, efficient criteria have been proposed in (\cite{c1}) to decide which components should be merged or split. Despite the fact that the number of Gaussian components are changing over time the shape of estimation distribution is not changed significantly since the set of data points is not increasing so fast. In the following the merge and split algorithms are briefly discussed:\\*
\\*
\noindent
\emph{Merge:}\\*
In order to reduce the number of components in a Gaussian distribution, some of the components should be merged together. However, an effective criterion is needed to pick the optimal pairs to merge.\\*
\textit{a) Criterion:} The posteriori probability of a data point gives a good estimation about which Gaussian component that data point belongs. If for many data points, the posteriori probabilities are almost equal for two different components, it can be perceived that the components are mergeable. To mathematically implement this, the following criterion for $ j $th and $ j^\prime $th Gaussian components is proposed in (\cite{c1}) $
J_{merge}(j,j^\prime;\hat{\lambda})=\textbf{P}_{j}(\hat{\lambda})^T \textbf {P}_{j^\prime}(\hat{\lambda})$
where $  \textbf {P}_{j}(\hat{\lambda})=(P(j | O_{1},\hat{\lambda}),P(j | O_{2},\hat{\lambda}),...,P(j | O_{t},\hat{\lambda}))^T $ is a $ N $-dimensional vector consisting of posteriori probabilities of all data points for $ j $th Gaussian component. The criterion $ J_{merge}(j,j^\prime;\hat{\lambda}) $ must be calculated for all possible pairs and the pair with the largest value is a candidate for the merge.\\*
\textit{b) Merging Procedure:} In order to merge two Gaussian components, the distribution model parameters must be re-estimated. A modified EM algorithm is proposed in (\cite{c1}) that re-estimates Gaussian parameters based on the former distribution parameters ($\hat{\lambda}$). If the merged Gaussian from the pair of $ j $ and $ j^\prime $ is denoted by $ j^{\prime\prime} $ then the initial parameters for the modified EM algorithm is $
{\omega^0_{j^{\prime\prime}}}=\omega_{j}+\omega_{ j^\prime},~
\mu^0_{j^{\prime\prime}}=\dfrac{\omega_{j} \mu_{j}+\omega_{ j^\prime} \mu_{ j^\prime}}{\omega_{j}+\omega_{ j^\prime}},~
\Sigma^0_{j^{\prime\prime}}=\dfrac{\omega_{j} \Sigma_{j}+\omega_{ j^\prime} \Sigma_{ j^\prime}}{\omega_{j}+\omega_{ j^\prime}}$. The initial parameter values calculated by the latter are often poor. Hence, the newly generated Gaussians should be
first processed by fixing the other Gaussians through the modified EM. An EM iterative algorithm then run to re-estimate the distribution parameters. The main steps are the same as (\ref{eq:11}) except the posteriori probability:
\begin{equation}
\label{eq:15}
\begin{split}
P(j^{\prime\prime} | O_{\tau},\hat{\lambda})=\dfrac{{\hat{\omega}}_{j^{\prime\prime}} ~g(O_{\tau}|{\hat{\mu}}_{j^{\prime\prime}},{\hat{\sigma}}_{j^{\prime\prime}})~\sum_{k=j,j^\prime} P(k | O_{\tau},\hat{\lambda})}{\sum_{k=j^{\prime\prime}} {\hat{\omega}}_{k} ~g(O_{\tau}|{\hat{\mu}}_{k},{\hat{\sigma}}_{k})}.
\end{split}
\end{equation}

By using this modified EM algorithm, the parameters of $ j^{\prime\prime} $th Gaussian are re-estimated  
without affecting the other Gaussian components. In our study, the merging algorithm could be repeated for several times if more than one merge step was needed.\\*
\\*
\noindent
\emph{Split:}\\*
In case of a need to increase the number of components in the estimation distribution, we use the split algorithm to split one or more Gaussians. As for the merging process, an appropriate criterion is necessary.\\*
\textit{a) Criterion:} As the split criterion of $ k $th component, the local Kullback-Leibler divergence is proposed in (\cite{c1}):
$
J_{split}(k;\hat{\lambda})=\int p_{k}(x,\hat{\lambda}) \log\allowbreak(\dfrac{p_{k}(x,\hat{\lambda})}{g(x|\hat{\mu_{k}},\hat{\Sigma_{k}})})~dx,
$
where $ p_{k}(x,\hat{\lambda}) $ is the local data density around $ k $th component\nomenclature{$J_{split}$}{split criterion for a Gaussian component} and is defined as $$
p_{k}(x,\hat{\lambda})=\dfrac{\sum_{n=1}^t~\delta(x-x_{n})P(k|x_{n},\hat{\lambda})}{\sum_{n=1}^tP(k|x_{n},\hat{\lambda})}$$. The split criterion $ J_{split}(k,\hat{\lambda}) $, which represents the distance between two Gaussian components, must be applied over all candidates and the one with the largest value will be selected.\\*
\textit{b) Splitting Procedure:} A modified EM algorithm is proposed in (\cite{c1}) to re-estimate the Gaussian parameters. If the split candidate is the $ k $th Gaussian component and the two resulting Gaussians are denoted by $ j^\prime $ and $ k^\prime $ the initial conditions are calculated as follows:
\begin{align}
\label{eq:18.n}
\begin{aligned}
\omega^0_{j^\prime}=\omega^0_{k^\prime}=\dfrac{1}{2} ~\omega_{k},~
\Sigma^0_{j^\prime}=\Sigma^0_{k^\prime}=\det(\Sigma_{k})^{1/d} ~I_{d},
\end{aligned}
\end{align}
where $ I_{d} $ is the $ d $-dimensional unit matrix and $ d $ is the dimension of Gaussian function $ g(x|\mu_{k},\Sigma_{k}) $. The mean vectors $ \mu^0_{j^\prime} $ and $ \mu^0_{k^\prime} $ are determined by applying random perturbation vector {\large $ \epsilon_{m} $}, $ m=1,2 $ on $ \mu_{k} $ as $ \mu^0_{j^\prime}=\mu_{k}+${\large $ \epsilon_{1} $} and $ \mu^0_{k^\prime}=\mu_{k}+${\large $ \epsilon_{2} $} where $ || ${\large $ \epsilon_{m} $}$||$$ \ll $$ || $$ \mu_{k}$$||$ and {\large $ \epsilon_{1} $}$ \neq ${\large $ \epsilon_{2} $}. The parameters re-estimation for $ j^{\prime} $ and $ k^\prime $ can be done by a modified EM algorithm similar to the merge EM algorithm where the modified posteriori probability is
\begin{equation}
\label{eq:19.n}
\begin{split}
P(m^\prime | O_{\tau},\hat{\lambda})=\dfrac{{\hat{\omega}}_{m^\prime} ~g(O_{\tau}|{\hat{\mu}}_{m\prime},{\hat{\sigma}}_{m^\prime})~\sum_{l=k} P(l | O_{\tau},\hat{\lambda})}{\sum_{l=j^\prime,k^\prime} \hat{\omega_{l}} ~g(O_{\tau}|\hat{\mu_{l}},\hat{\sigma_{l}})},
\end{split}
\end{equation}
where $ m^\prime=j^\prime, k^\prime $. \nomenclature{$J_{com}$}{communication cost} The parameters of $ j^{\prime} $ and $ k^\prime $ are re-estimated without affecting other Gaussians. Splitting algorithm will be repeated if more than one split was necessary according to the Akaike criterion. By means of AIC and split-merge technique, the agents are able to estimate the number of targets and estimate the $GMM$ without knowing the number of agents.
\begin{figure}[!t]
	\centering
	\subfloat[The logic behind the revision probability]{{\includegraphics[width=0.69\textwidth]{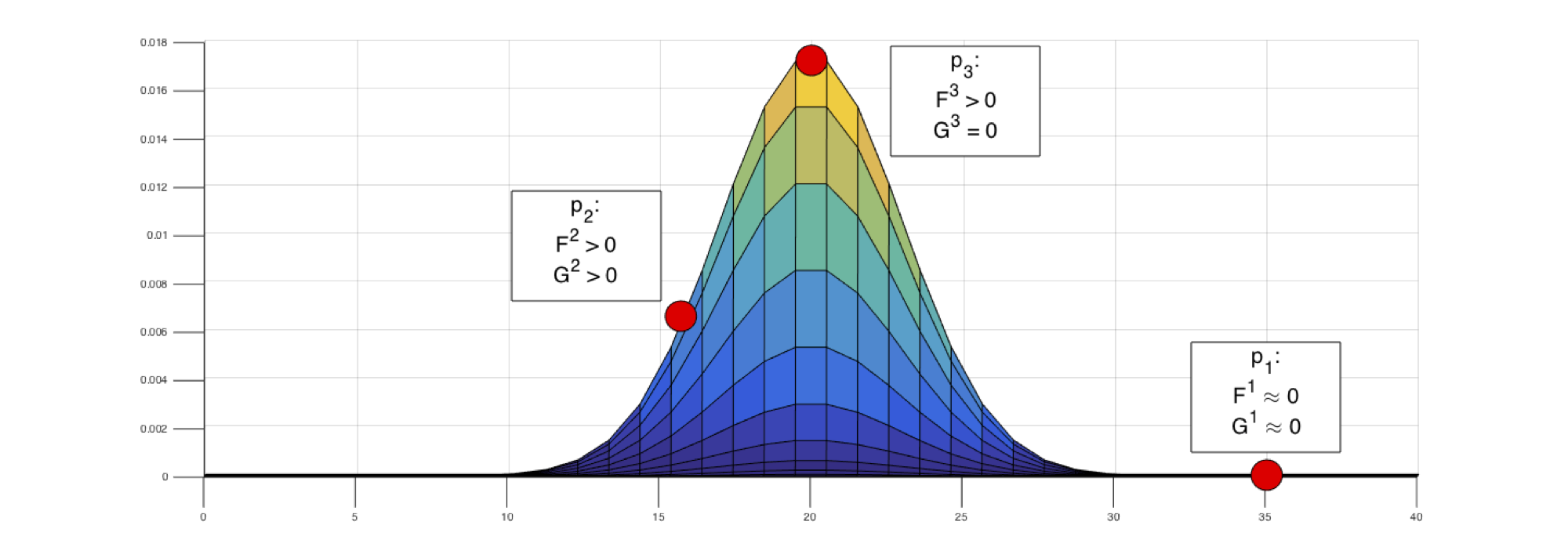}}}
	~
	\subfloat[Revision probability function]{{\includegraphics[width=0.3\textwidth]{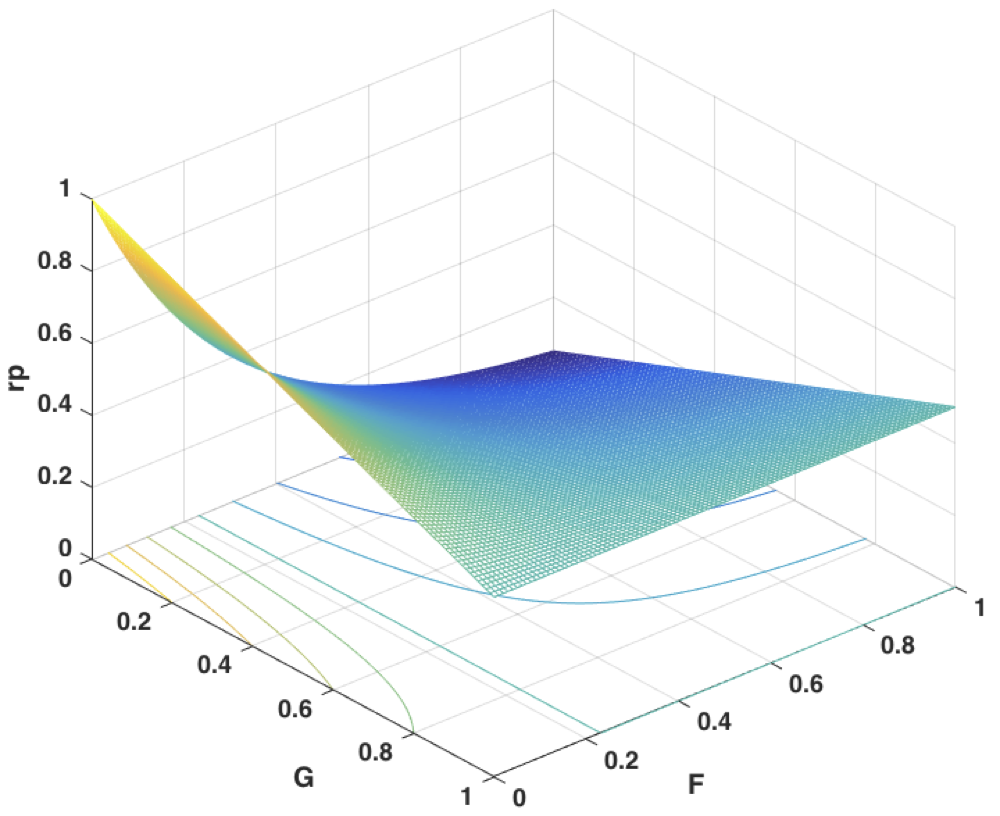}}}
	\caption{Revision Probability}
	\label{f2}
\end{figure}
\\*
\\*
\noindent
\emph{Revision Probability:}
\\*
Recall that the revision probability $rp^i$ in P-SBLLL is the probability with which agent $i$ wakes up to update its action. We considered a two variable function for the revision probability of each agent. The corresponding probability of each player's revision probability depends on the player's action, i.e. player's location in our MCC example. Each player $i$ can determine the outcome of his action as the signal strength $f^i(l) $ of his location $\alpha^i=l$. Furthermore, we assume that each agent can sense a local gradient of the worth denoted by $ g^i(l) $. At each iteration, $ f^i(l) $ and $ g^i(l) $ are normalized based on the player's maximum observed $ f^i(l) $ and $ g^i(l) $. For each agent $i$, let $ F^i $ and $G^i$ be the normalized versions of $ f^i(l) $  and $ g^i(l) $ respectively.

We desire a revision probability function ($rp^i(F^i,G^i)$) for which the probability that players wake up depends on each player's situation.We consider three situations which a robot can fall in (Fig.\ref{f2}.a): 
\begin{itemize}
	\item \textit{Situation $p_1$ where player's received signal and gradient is almost zero:} In such situation, player has to wake up to update its action, i.e. explore, and to move toward worthwhile areas. Let $a_1$ be the probability that player in $ p_1 $ wakes to update its action.
	\item \textit{Situation $p_2$ where player's received signal and gradient is relatively high:} In this situation, player has entered a worthwhile region. Since the player already entered a worthwhile area, the need for exploration reduces comparing to the player in $ p_1 $, i.e., $a_2<a_1$ where $a_2$ is the probability that player in $ p_2 $ wakes up.
	\item \textit{Situation $p_3$ where player's received signal is relatively high but the gradient is almost zero:} This situation happens when player reaches a high local value. Clearly, player in $p_3$ has to remain on its position, i.e. remain asleep. Hence, $a_3<a_2<a_1$ where $a_3$ is the probability that player in $ p_3 $ wakes up.
\end{itemize}

To match with the shape of Gaussian function, we prefer to have an exponential drop as player $i$'s normalized signal strength $F^i$ increases. Hence, when $G^i=0$, the form of revision probability function is as follow:
$$
rp^i(F^i,G^i=0)=e^{-k(F^i-c)},
$$
where $k$ is the drop rate and $c:=\dfrac{\ln (a_1)}{k}$. For the sake of simplicity, we consider a linear change in revision probability function as $G$ increases. As $G$ reaches 1, the value of the revision probability function must be equal to $a_2$. Therefore, for a constant $F$, the slope of the line from $G=0$ to $G=1$ is $a_2-e^{-k(F^i-c)}$. Furthermore, y-intercept for this line is $e^{-k(F^i-c)}$. Thus the complete function is of the form

\begin{equation}
\label{rpi}
rp^i(F^i,G^i)=(a_2-e^{-k(F^i-c)})G+e^{-k(F^i-c)}.
\end{equation}

As in Fig. \ref{f2}.b, the function behavior is linear versus $ G $ and exponential versus $ F $. With this independent revision, each player decides based on its status whether it is a good time to wake up or not. We believe it is more efficient than a random selection as in BLLL. Furthermore, it reduces the need for unnecessary trial and errors.
\\*
\\*
\noindent
\emph{Simulation Results:}
\\*
Assume that all agents adhere to P-SBLLL and Assumptions \ref{ass2} and \ref{ass3} hold for the MCC problem. Players' utility function is separable and from Proposition \ref{prop1}, we know that the stochastically stable states are the set of potential maximizers. The simulation parameters are chosen as $ K^{i}=3 \times 10^{-5} $ for $ i=1,...,N $, $ \delta=1.5 $, $ R_{com}=56 $, $a_1=1$, $a_2=0.5$ and $a_3=0.1$.

\begin{figure}[!t]\label{sblllpsfig}
	set $ n=1 $\\
	\textbf{for} each robot $i \in \mathcal{I}$ \textbf{do}\\
	initialize $\alpha^i(1) \in \mathcal{L}$ randomly\\
	\textbf{while} the covered worth $\sum_{i \in \mathcal{I}} C^i(\alpha^i)$ is not in a steady state \textbf{do}\\
	$ ~~~~~~ $ \textbf{if} $T_{AIC}$ is a sub-multiple of $n$ \textbf{then}\\
	$ ~~~~~~ $ $ ~~~~~~ $ process merge or split\\
	$ ~~~~~~ $ \textbf{for} each robot $i \in \mathcal{I}$ \textbf{do}\\
	$ ~~~~~~ $ $ ~~~~~~ $ determine $rp^i$ (see (\ref{rpi}))\\
	$ ~~~~~~ $ $ ~~~~~~ $ robot $i$ wakes up with probability $rp^i$ \\
	$ ~~~~~~ $ $ ~~~~~~ $ (let $S(n)$ be the set of robots that are awake at $n$)\\
	$ ~~~~~~ $ \textbf{for} each robot $i \in S(n)$ \textbf{do}\\
	$ ~~~~~~ $ $ ~~~~~~ $ robot $i$ chooses a trial action $\alpha^i_T$ randomly from $\mathcal{A}^i_c$\\
	$ ~~~~~~ $ $ ~~~~~~ $ calculate $X^i_{\alpha^i(n)}$ and $X^i_{\alpha^i_T}$ by using both $GMM$ and $\widehat{GMM}$ (see (\ref{eq:12.8.4n}) and (\ref{eq:12.8.5n}))\\
	$ ~~~~~~ $ $ ~~~~~~ $ $\alpha^i(n+1)\leftarrow \alpha^i(n)$ with probability $X^i_{\alpha^i(n)}$\\
	$ ~~~~~~ $ $ ~~~~~~ $ or\\
	$ ~~~~~~ $ $ ~~~~~~ $ $\alpha^i(n+1)\leftarrow \alpha^i_T$ with probability $X^i_{\alpha^i_T}$\\
	$ ~~~~~~ $ $ ~~~~~~ $ \textbf{if} $\alpha^i(n+1)$ is $ \alpha^i_T$ \textbf{then}\\
	$ ~~~~~~ $ $ ~~~~~~ $ $ ~~~~~~ $ add $\alpha^i(n+1)$ to $O^i$\\
	$ ~~~~~~ $ $ ~~~~~~ $ $ ~~~~~~ $ run EM and update $\widehat{GMM}$\\
	$ ~~~~~~ $ \textbf{for} each robot $j \in \mathcal{I} \setminus S(n)$ \textbf{do}\\
	$ ~~~~~~ $ $ ~~~~~~ $ $\alpha^{j}(n+1)\leftarrow \alpha^{j}(n)$\\
	$ ~~~~~~ $ $n \leftarrow n+1$
	\caption{P-SBLLL's pseudo code}
\end{figure}

\begin{figure}[!t] \centering \includegraphics[width=0.7\textwidth]{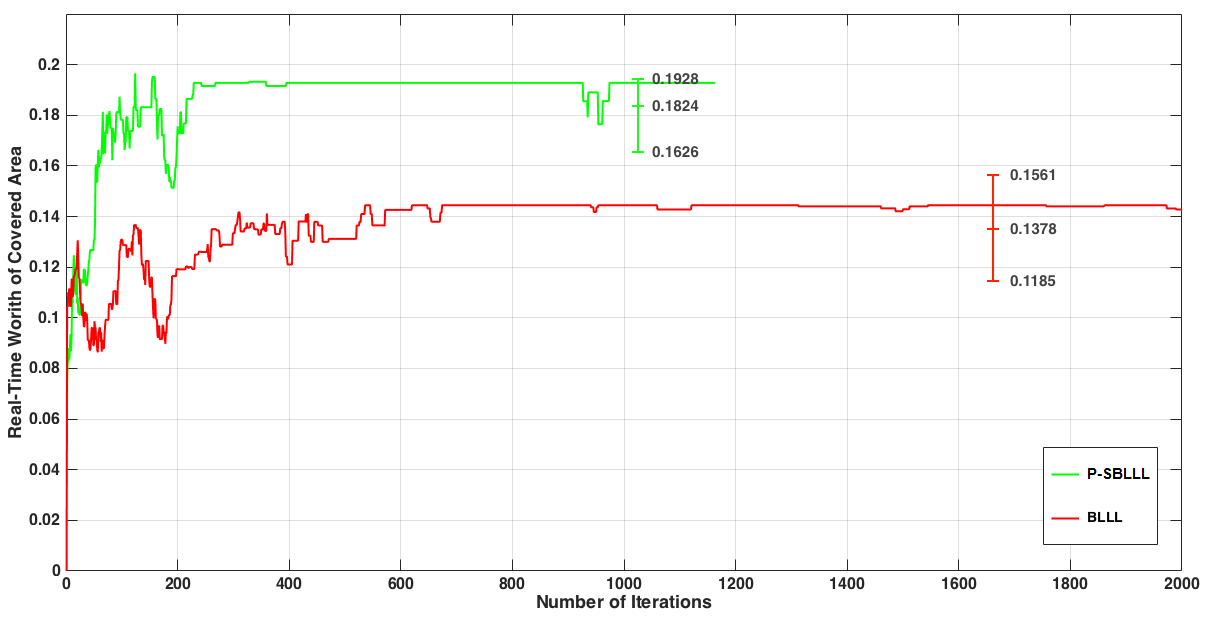} \caption{The real-time worth of the covered area}
	\label{f1} 
\end{figure}
\begin{figure}[!t] \centering \includegraphics[width=0.7\textwidth]{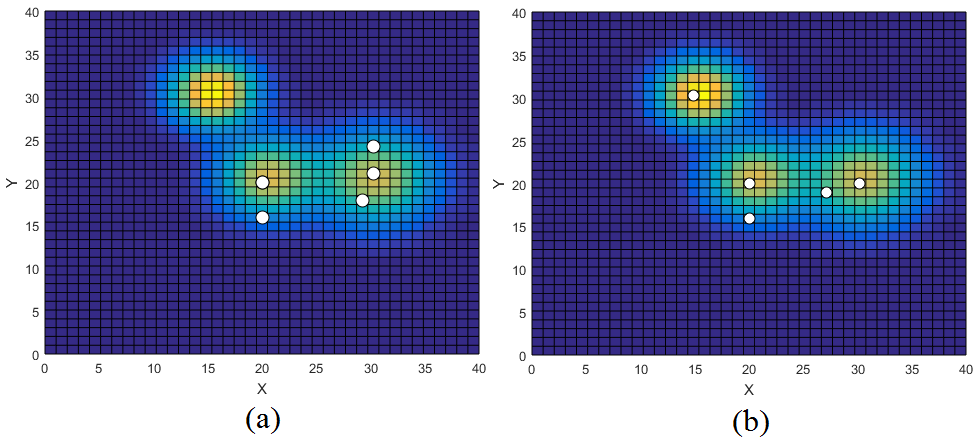} \caption{Final configuration of agents in (a) BLLL and (b) P-SBLLL}
	\label{SBLLL_FC} 
\end{figure}
\begin{figure}[!t]
	set $ n=1 $\\
	\textbf{for} each robot $i \in \mathcal{I}$ \textbf{do}\\
	initialize $\alpha^i(1) \in \mathcal{L}$ randomly\\
	initialize $Q^i(0)$ and $Q^i(1) \in {\mathds{R}}^{|\mathcal{L}|}$\\
	initialize $X^i \in {\mathds{R}}^{|\mathcal{L}|}$\\
	\textbf{while} the covered worth $\sum_{i \in \mathcal{I}} C^i(\alpha^i)$ is not in a steady state \textbf{do}\\
	$ ~~~~~~ $ \textbf{for} each player $i \in \mathcal{I}$ \textbf{do}\\
	$ ~~~~~~ $ $ ~~~~~~ $ perform an action $\alpha^i(n)=\beta^i$ from $\mathcal{A}^i_c$ based on $X^i(n)$\\
	$ ~~~~~~ $ $ ~~~~~~ $ receive $u^i(n)$\\
	$ ~~~~~~ $ $ ~~~~~~ $ $u_\beta^i(n) \leftarrow \mathds{1}_{\{\alpha^i(n)=\beta\}}~
	u^i(n)$\\
	$ ~~~~~~ $ $ ~~~~~~ $ $Q_\beta^i(n+1)\leftarrow 2(1-\mu^i) Q_\beta^i(n)-(1-\mu^i)^2 Q_\beta^i(n-1)+{\mu^i}^2 u_\beta^i(n)$\\
	$ ~~~~~~ $ $ ~~~~~~ $ \textbf{for} each $\beta ' \in \mathcal{A}^i, \beta ' \neq \beta$ \textbf{do}\\
	$ ~~~~~~ $ $ ~~~~~~ $ $ ~~~~~~ $ $Q_{\beta '}^i(n+1)\leftarrow Q_{\beta '}^i(n)$\\
	$ ~~~~~~ $ $ ~~~~~~ $ $X^i(n+1)\leftarrow (1-\vartheta)X^i(n)+\vartheta BR^i(Q^i(n))$\\
	$ ~~~~~~ $ $n \leftarrow n+1$
	\caption{SOQL's pseudo code}
\end{figure}
\begin{figure}[!t] \centering \includegraphics[width=0.7\textwidth]{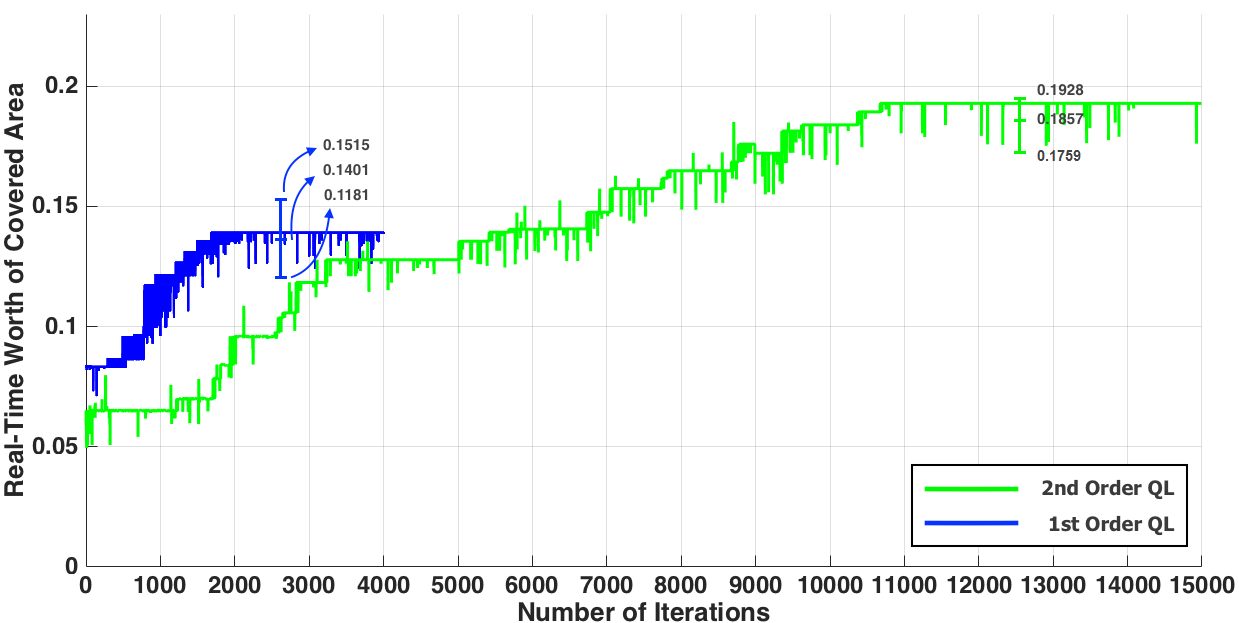} \caption{The real-time worth of the covered area}
	\label{f4nn}
\end{figure} 
\begin{figure}[!t] \centering \includegraphics[width=0.7\textwidth]{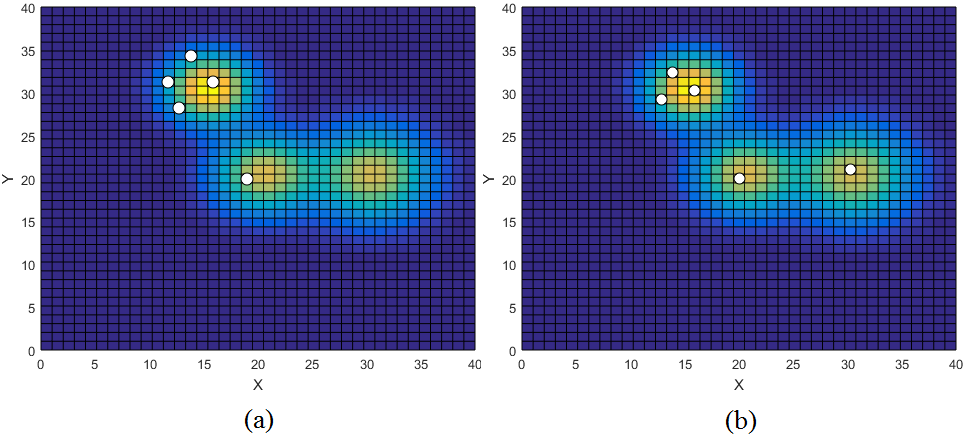} \caption{Final configuration of agents in (a) standard Q-learning and (b) SOQL}
	\label{1QL_FC}
\end{figure}
\begin{figure}[!t] \centering \includegraphics[width=0.7\textwidth]{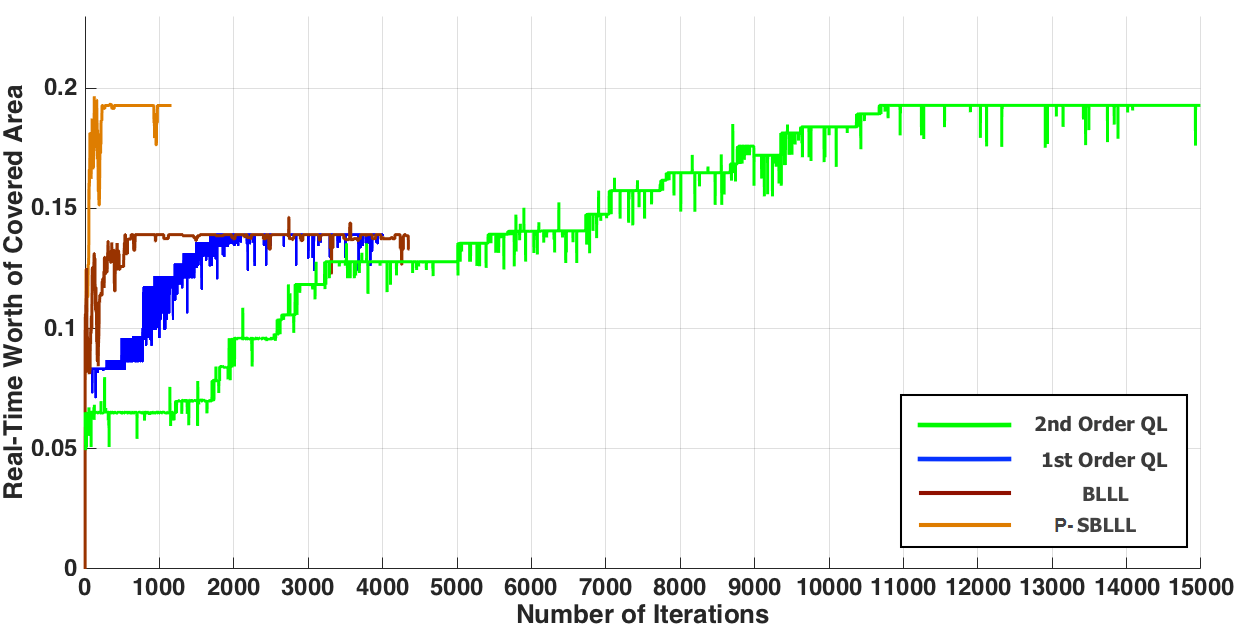} \caption{The real-time worth of the covered area}
	\label{f4nnn}
\end{figure}

The worth of covered area using P-SBLLL and BLLL is shown in Fig. \ref{f1}. Clearly, the covered worth is higher and the convergence rate is faster in P-SBLLL. Comparing to BLLL, in P-SBLLL each player autonomously decides, based on its situation, to update its action while in BLLL only one random player is allowed to do the trial and error. The algorithms ran for 10 times with different initial conditions and for each algorithm Fig. \ref{f1} presents a bound and a mean for the coverage worth. The final configuration of the agents are shown in Fig. \ref{SBLLL_FC}. We can see that in P-SBLLL the agents found all the targets. Although three agents have the targets in their sensing range, the two other agents tried to optimize their configuration with respect to the signal distribution to maximize the coverage worth.

\subsubsection{SOQL}
In SOQL, the robots sample the environment at each time to create a memory of the payoff of the actions that they played. When the environment is explored enough, this memory can help the robots to find the game's global optimum. 

The simulation parameters are chosen as $ K^{i}=3 \times 10^{-5} $ for $ i=1,...,N $, $ \delta=1.5 $, $\mu=0.97$, $\vartheta=0.5$, $\zeta=0.9999$ and $\xi=0.01$. The worth of covered area by all robots, using SOQL and first-order Q-learning, is shown in Fig. \ref{f4nn}. It can be seen that the convergence rate is lower for the SOQL algorithm comparing to the first-order algorithm. However, the covered worth is higher comparing to the first-order case. The algorithms ran for 5 times with different initial conditions and for each algorithm Fig. \ref{f4nn} presents a bound and a mean for the coverage worth.

Fig. \ref{1QL_FC} shows the final configuration of the robots in standard QL and SOQL. It can be seen that the robots that used SOQL as their learning algorithm, successfully found all the targets. However, with the same initial locations the robots who used standard QL could not find all the targets.

\section{Conclusion}
By relaxing both asynchrony and completeness assumptions P-SBLLL in Section \ref{sblll} enhanced the way the group of agents interact with their environment. We showed that comparing to BLLL, P-SBLLL's performance is excellent in a model-based learning scheme. A higher convergence rate in P-SBLLL, as it was expected, is because players learn in parallel. Furthermore, because of this simultaneous learning in P-SBLLL, agents can widely explore the environment. Another valuable feature of P-SBLLL is that each agent's exploration is dependent on the agent's situation in the environment. Thus, each agent can autonomously decide whether it is a good time for exploration or it is better to remain on its current state. This will certainly reduce redundant explorations in P-SBLLL comparing to BLLL.
	
In Section \ref{soql} we proposed SOQL, a second-order reinforcement to increase RL's aggregation depth. Comparing to the model-based P-SBLLL and BLLL, the convergence rate of SOQL is lower (Fig.\ref{f4nnn}) due to the need for a wide exploration in an RL scheme. However, in SOQL algorithm, the equilibrium worth is the same as P-SBLLL's equilibrium worth, and can be employed when the utility distribution is not a $GMM$ or more generally, when the utility structure is unknown.

\acks{This work was done while the first author was at University of Toronto. We would like to acknowledge support for this project from the Natural Sciences and Engineering Research Council of Canada (NSERC).}

\bibliographystyle{apalike}
\bibliography{template}
\end{document}